\pdfoutput=1

\documentclass[11pt, dvipsnames]{article}

\usepackage[preprint]{acl}
\usepackage{multirow}
\usepackage{kotex}

\usepackage{times}
\usepackage{latexsym}

\usepackage[T1]{fontenc}

\usepackage[utf8]{inputenc}

\usepackage{microtype}

\usepackage{inconsolata}

\usepackage{graphicx}

\usepackage{xcolor,colortbl}
\usepackage{subfigure}
\usepackage{booktabs}
\usepackage{arydshln}
\usepackage{array}

\usepackage{xspace}
\newcommand{\exaone}{{\sc Exaone}\xspace}

\title{ixi-GEN: Efficient Industrial sLLMs \\ through Domain Adaptive Continual Pretraining}

\author{
Seonwu Kim,~~Yohan Na,~~Kihun Kim,~~Hanhee Cho,~~Geun Lim, \textbf{Mintae Kim},\\
~~\textbf{Seongik Park},~~\textbf{Ki Hyun Kim}\thanks{Corresponding author.},~~\textbf{Youngsub Han},~~\textbf{Byoung-Ki Jeon}
\smallskip 
\\
LG Uplus \\
\{seonwukim, nayohan, kihunkim, hanhee, glim, iammt, spark32, \\
kimkihyun,  yshan042, bkjeon\}@lguplus.co.kr\\
}

\begin{document}
\maketitle
\begin{abstract}
The emergence of open-source large language models (LLMs) has expanded opportunities for enterprise applications; however, many organizations still lack the infrastructure to deploy and maintain large-scale models. As a result, small LLMs (sLLMs) have become a practical alternative despite inherent performance limitations. While Domain Adaptive Continual Pretraining (DACP) has been explored for domain adaptation, its utility in commercial settings remains under-examined. In this study, we validate the effectiveness of a DACP-based recipe across diverse foundation models and service domains, producing DACP-applied sLLMs (\emph{ixi-GEN}). Through extensive experiments and real-world evaluations, we demonstrate that \emph{ixi-GEN} models achieve substantial gains in target-domain performance while preserving general capabilities, offering a cost-efficient and scalable solution for enterprise-level deployment.

\end{abstract}

\section{Introduction}

The development of large language models (LLMs) was initially dominated by a few companies that restricted model weight parameters, but the release of LLaMA \cite{touvron2023llamaopenefficientfoundation} expanded LLM accessibility, narrowing the gap between proprietary and open models \cite{grattafiori2024llama3herdmodels,deepseekai2025deepseekr1incentivizingreasoningcapability,gemma2025gemma3}, leading many companies to focus on fine-tuning and continual learning rather than developing models from scratch \cite{singhal2022largelanguagemodelsencode,yang2023fingptopensourcefinanciallarge}. To address the performance limitations of small LLMs (sLLMs) in industrial applications \cite{li2024smalllanguagemodelsapplication,lu2025smalllanguagemodelssurvey}, this study proposes a continual pretraining methodology that optimizes sLLMs within service domains, improving performance and cost efficiency.

\begin{figure}[t]
  \centering
  \vspace{1mm}
  \includegraphics[width=\linewidth]{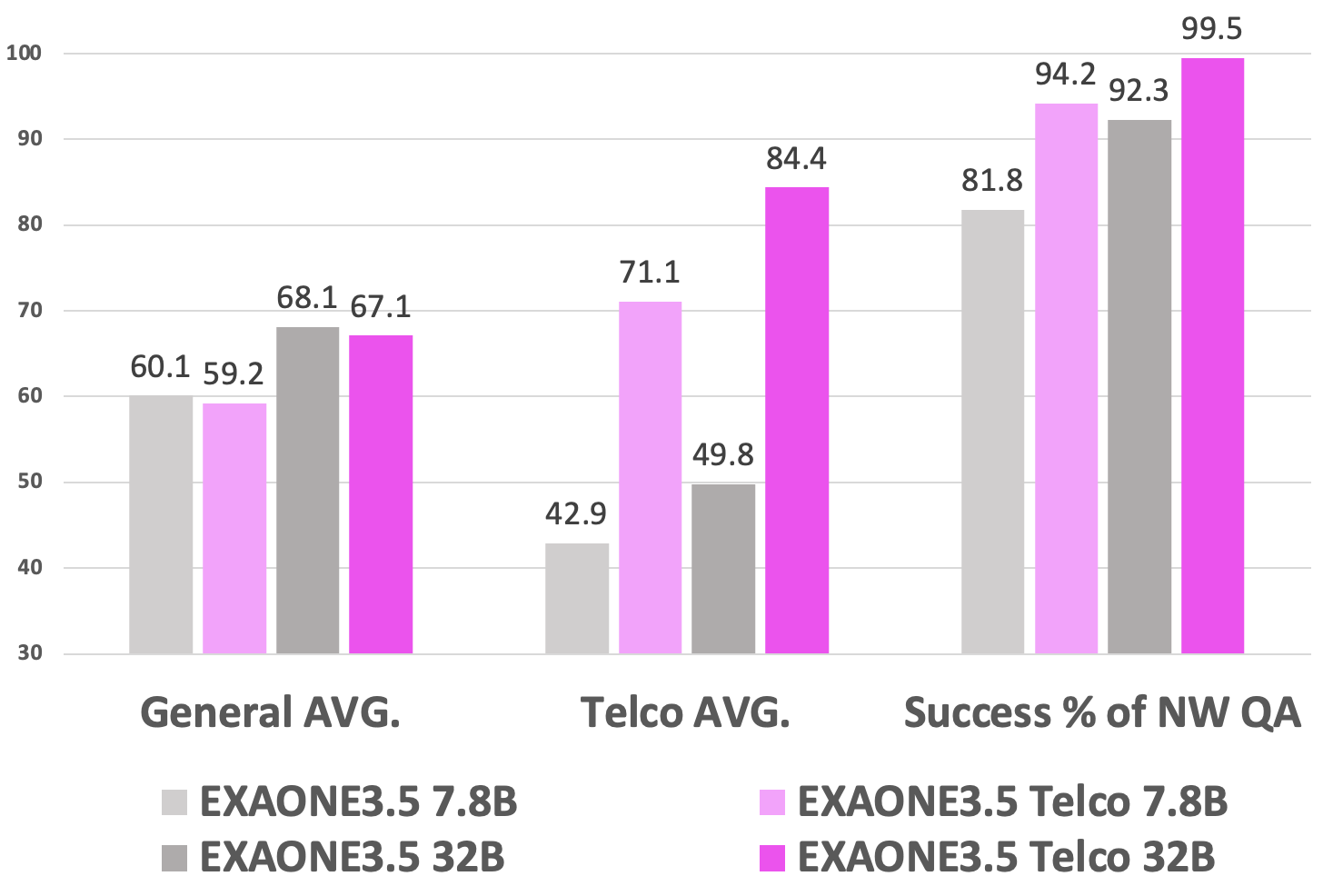}
  \caption{DACP-applied sLLM outperforms the base model in both the Telco domain and commercial service tasks, while preserving general domain capabilities, thereby enabling cost-efficient yet high-performance service delivery.}
  \vspace{-4mm}
  \label{fig:introduction_graph}
\end{figure}


\begin{figure*}[t]
  \centering
  \vspace{-3mm}
  \includegraphics[width=\linewidth]{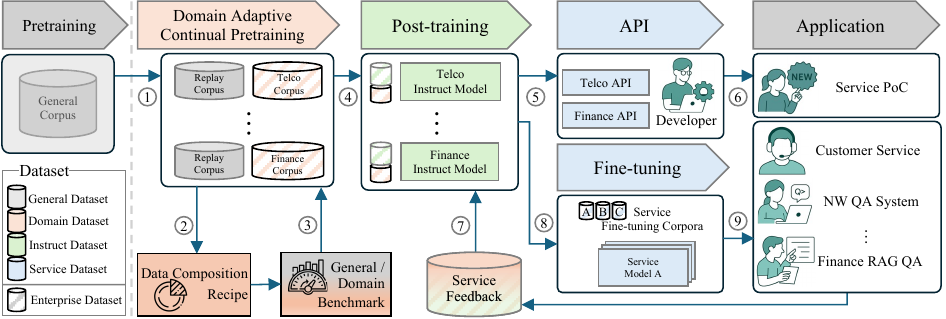}
  \caption{Workflow for domain-adapted sLLM development. The process begins with domain-adaptive continual pretraining (DACP) using both general and domain corpora, followed by instruction tuning to produce domain adapted models. These models are exposed via APIs and optionally fine-tuned, enabling commercial service applications. Evaluation and feedback loops ensure continuous improvement.}
  \vspace{-3mm}
  \label{fig:overall_process}
\end{figure*}


This study introduces Domain Adaptive Continual Pretraining (DACP) as a methodology to mitigate performance degradation in sLLMs and enhance their effectiveness in target domains. DACP offers an alternative to training models from scratch by continual pretraining a general-purpose LLM on a corpus of domain-specific unlabeled data \cite{xie2023efficientcontinualpretrainingbuilding}. Through experiments, we demonstrate that our method enables efficient domain adaptation across various domains, backbone models, and model sizes, and we further validate its practical viability through a case study on its successful integration into an industrial application. The overall process is illustrated in Figure \ref{fig:overall_process}.

The contributions of this study are as follows.




First, we demonstrate that DACP-applied smaller LLMs (sLLMs) can outperform larger general domain models within target domains, thereby significantly mitigating performance limitations due to model size. This finding highlights the potential for deploying high-performing yet cost-efficient sLLMs in real-world enterprise services.



Second, we conduct extensive experiments across multiple domains (Telco and Finance) and foundation models (LLaMA, Qwen, \exaone) to establish a robust DACP training recipe. Through ablation studies, we identify key factors contributing to performance gains and provide practical guidelines for domain adaptation.



Finally, beyond benchmark evaluations, we conduct human evaluations on real-world service tasks, such as customer service summarization and Telco/Finance QA, demonstrating that DACP-applied models deliver not only quantitative gains but also measurable improvements in user experience.

\section{Related Work}


Large language models (LLMs) are increasingly applied across various domains, with research exploring diverse tasks that leverage their embedded knowledge \cite{kaur2024texttransformationcomprehensivereview}. However, as models become outdated, retraining from scratch is prohibitively expensive, highlighting the need for continual learning, which updates models using newly added data rather than an entirely new dataset \cite{jang2023temporalwikilifelongbenchmarktraining,wu2024continuallearninglargelanguage,vandeven2019scenarioscontinuallearning,parisi2019continual}.


The need for continual learning has been emphasized to address the limitations of open models, which were initially developed with an English-centric focus and exhibited lower performance in other languages \cite{wu2024continuallearninglargelanguage,zhao2024largelanguagemodelshandle}. While advancements in open models have helped alleviate language imbalance \cite{zhao2024llamaenglishempiricalstudy,cui2024efficienteffectivetextencoding,kim2024efficienteffectivevocabularyexpansion}, researchers continue to explore domain adaptation methods to further enhance the model \cite{ling2024domainspecializationkeymake,song2025injectingdomainspecificknowledgelarge,lu2024finetuninglargelanguagemodels}.


Efforts to enhance model adaptation have ranged from fine-tuning small-scale corpora to continual pretraining on medium-scale corpora, typically involving hundreds of gigabytes of data \cite{gupta2023continualpretraininglargelanguage}. However, continual pretraining requires frequent model updates, which can cause catastrophic forgetting, a persistent challenge in continual learning research \cite{parisi2019continual}, that we address through mitigation strategies and experimentally verify to maintain general performance.



Our method should be applied to the foundation model before instruction tuning \cite{gururangan2020dontstoppretrainingadapt}, necessitating post-training to restore instruction-following capabilities. Fortunately, this topic has been actively studied in the open-source community \cite{sanh2021multitask}. 

Unlike previous work \cite{jiang2024improving,wu2023bloomberggpt}, primarily focused on domain adaptation itself, we extend LLM domain adaptation research by verifying an adaptation recipe across various domains and backbone models, while also evaluating its effectiveness and industrial efficiency.


\section{DACP Recipe}



In industry, domain adaptation of LLMs has traditionally relied on fine-tuning smaller domain datasets \cite{ling2024domainspecializationkeymake}. However, supervised fine-tuning (SFT) primarily leverages pretrained knowledge rather than systematically incorporating new domain knowledge \cite{gekhman-etal-2024-does}. To address this, we propose leveraging mid-scale domain corpora during continual pretraining to enhance domain adaptation. This approach, however, involves repeated updates to the weight parameters and thus introduces the risk of catastrophic forgetting of pretrained knowledge \cite{yogatama2019learning,liu-etal-2024-catastrophic}.

This phenomenon poses a significant challenge in LLM domain adaptation: overly aggressive adaptation can erode general knowledge and reasoning ability. Consequently, maintaining an appropriate balance between domain expertise and pretrained knowledge is crucial for achieving high performance on domain tasks while preserving reliable responses to general queries \cite{li2017learningforgetting,liao2024exploringforgettinglargelanguage}, as illustrated in Figure~\ref{fig:why_we_need_dacp}.

During the DACP process, training is conducted using a mid-scale corpus. However, a key concern is the risk of catastrophic forgetting, where knowledge acquired during pretraining may be lost while learning new domain knowledge \cite{yogatama2019learning}. This phenomenon can lead to a decrease in overall model performance, underscoring the need for effective mitigation strategies.

\begin{figure}[t]
  \centering
  \includegraphics[width=\linewidth]{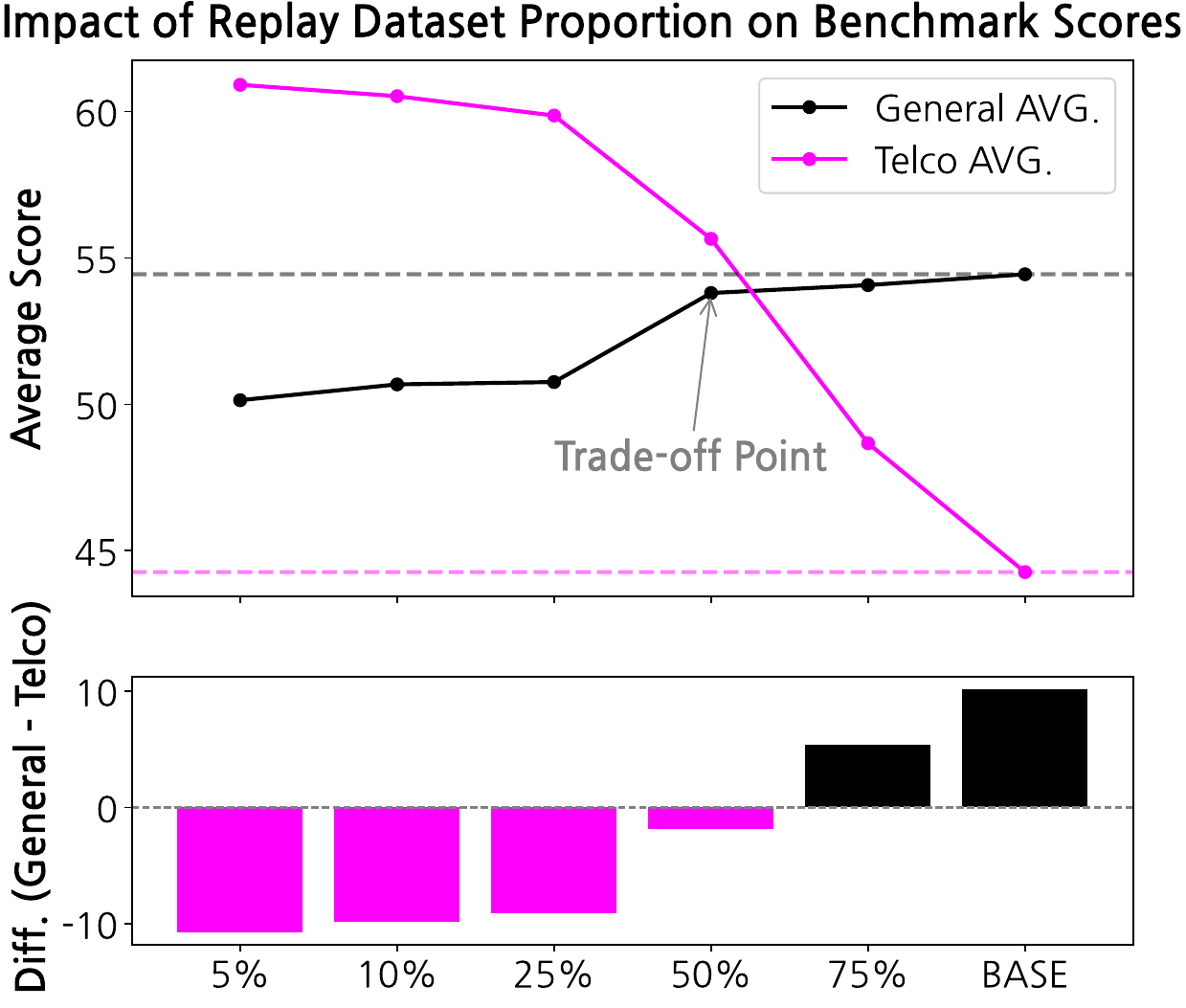}
  \caption{Model performance on general and Telco benchmarks with varying proportions of replay and Telco dataset proportions.}
  \vspace{-4mm}
  \label{fig:telco_dacp_ratio}
\end{figure}

To mitigate this risk, we adopt replay datasets, which have been shown to be effective in continual pretraining \cite{scialom2022finetunedlanguagemodelscontinual}. Specifically, we construct a pseudo-replay dataset from widely used public corpora—FineWeb \cite{fineweb2024}, CC \cite{commoncrawl2025}, Wikipedia, and GitHub Code—given that most open-weight models do not disclose their dataset compositions. To further improve Korean language performance, we additionally incorporate substantial Korean corpora from AIHub \cite{aihub2025} and NIKL \cite{klri2025}.


To balance general-domain retention and domain learning, we conducted a preliminary study on the replay ratio using the \exaone-3.5 2.4B model and a 3B token dataset in the Telco domain. Varying the replay data ratio, we evaluated performance on general-domain \cite{son2024kmmlumeasuringmassivemultitask,son2024haeraebenchevaluationkorean} and Telco benchmarks. As shown in Figure~\ref{fig:telco_dacp_ratio}, a 50\% replay ratio effectively preserves general capabilities while improving domain performance. We therefore adopted this ratio for the full DACP corpus (Table~\ref{tab:telcoDataRatio}); further details are in Appendix~\ref{sec:replay_ratio_selection_detail}.


\section{Data Preparation}

\subsection{DACP Dataset Preparation}


We constructed a training dataset by combining a target domain dataset with a pseudo-replay dataset. The Telco dataset is primarily composed of Telco customer service conversation transcriptions from speech recognition, supplemented with telecommunication and network-related knowledge. Due to typographical errors from speech recognition in this pseudo corpus, we developed an error correction model to improve pretraining efficiency. In Appendix \ref{sec:telco_preprocessing} for details of the preprocessing and error correction pipeline. To promote deeper contextual understanding, a subset of data was annotated with supporting information. The final dataset composition is shown in Table \ref{tab:telcoDataRatio}. Meanwhile, the financial domain dataset was also constructed, with details summarized in the Appendix \ref{sec:finance_dacp_datasets}.

\begin{table}[h] \label{tab:dataRatio}
    \footnotesize
    \centering
    \begin{tabular}{
        >{\raggedright\arraybackslash}p{0.2\textwidth}
        >{\centering\arraybackslash}p{0.1\textwidth}
        >{\centering\arraybackslash}p{0.1\textwidth}
    }
        \toprule 
        \textbf{Dataset} &\textbf{Size} &\textbf{Ratio} \\
        \midrule
        Telco Corpora &140GB &45.90\% \\
        Korean Corpora &100GB &32.97\% \\
        English Corpora &50GB &16.18\% \\
        Code &15GB &4.92\% \\
        \midrule
        \textbf{Total} &305GB &100\% \\
        \bottomrule
    \end{tabular}
    \vspace{-2mm}
    \caption{Composition of training-set for Telco DACP.}
    \vspace{-4mm}
    \label{tab:telcoDataRatio}
\end{table}

\begin{table*}[hbt!]
    \centering
    \footnotesize 
    \begin{tabular}{
        >{\raggedright\arraybackslash}p{0.22\textwidth}
        >{\centering\arraybackslash}p{0.09\textwidth}
        >{\centering\arraybackslash}p{0.09\textwidth}
        >{\centering\arraybackslash}p{0.10\textwidth}
        >{\centering\arraybackslash}p{0.10\textwidth}
        >{\centering\arraybackslash}p{0.10\textwidth}
        >{\centering\arraybackslash}p{0.11\textwidth}
    }
    \toprule  
    \multicolumn{7}{c}{\textbf{General Domain Benchmark}} \\ 
    \midrule
    \textbf{Model} & \textbf{MMLU} & \textbf{BBH} & \textbf{KMMLU} & \textbf{HAERAE} & \textbf{GSM8k-Ko} & \textbf{Avg.} \\
    \midrule
    \multicolumn{7}{l}{\textbf{Original Instruction Model (Vanilla)}} \\
    Llama 3.2 3B IT  & 59.77 & 57.10 & 34.23 & 42.44 & 33.59 & 45.43 \\ 
    Qwen 2.5 3B IT  & 66.48 & 34.20 & 34.32 & 62.42 & 44.50 & 48.38 \\ 
    \exaone 3.5 2.4B IT  & 59.30 & 47.93 & 42.77 & 66.18 & 47.01 & 52.64 \\ 

    \hdashline
    Llama 3.1 8B IT  & 68.07 & 71.03 & 41.66 & 63.79 & 52.62 & 59.43 \\
    Qwen 2.5 7B IT  & 74.21 & 57.69 & 46.90 & 75.80 & 60.27 & 62.97 \\ 
    \exaone 3.5 7.8B IT  & 65.35 & 56.97 & 44.93 & 78.28 & 54.97 & 61.01 \\
    \hdashline
    \exaone 3.5 32B IT  & 74.04 & 67.79 & 51.12 & 83.50 & 64.29 & 68.15 \\
    
    \midrule
    \multicolumn{7}{l}{\textbf{Telco DACP Model (Ours)}} \\ 
    Llama 3.2 Telco 3B IT       & 52.97 & 30.15 & \textbf{36.49} & \textbf{57.93} & 33.28 & 42.16 (-7\%) \\

    Qwen 2.5 Telco 3B IT        & 64.06 & 18.61 & \textbf{46.38} & \textbf{62.97} & \textbf{50.80} & \textbf{48.56 (+0\%)} \\
    \exaone 3.5 Telco 2.4B IT   & 54.89 & 32.31 & 37.37 & \textbf{66.64} & 38.97 & 50.03 (-5\%) \\
    \hdashline
    Llama 3.1 Telco 8B IT       & 63.74 & \textbf{71.66} & \textbf{41.77} & \textbf{70.21} & 45.56 & 58.59 (-1\%) \\
    Qwen 2.5 Telco 7B IT        & 71.16 & \textbf{61.50} & \textbf{53.20} & 70.76 & \textbf{62.93} & \textbf{63.91 (+1\%)} \\
    \exaone 3.5 Telco 7.8B IT   & 63.19 & \textbf{59.30} & 43.70 & 74.70 & \textbf{55.19} & 59.22 (-1\%) \\
    \hdashline
    \exaone 3.5 Telco 32B IT    & 70.61 & \textbf{70.68} & 47.05 & 82.13 & \textbf{64.90} & 67.07 (-2\%) \\
    
    \bottomrule
    \end{tabular}
    \vspace{-1mm}
    \caption{Performance comparison of original instruction models and Telco Domain Adapted models (Telco DACP) evaluated on the General Domain Benchmark. Models are grouped by size (dashed lines). Percentage improvements over original models are shown in parentheses. \textbf{Bold} scores indicate an improvement over the instruction model.} 
    \vspace{-2mm}
    \label{tab:generalBench}
\end{table*}

\begin{table*}[hbt!]
    \centering
    \footnotesize  
    \begin{tabular}{
        >{\raggedright\arraybackslash}p{0.22\textwidth}
        >{\centering\arraybackslash}p{0.09\textwidth}
        >{\centering\arraybackslash}p{0.09\textwidth}
        >{\centering\arraybackslash}p{0.11\textwidth}
        >{\centering\arraybackslash}p{0.08\textwidth}
        >{\centering\arraybackslash}p{0.1\textwidth}
        >{\centering\arraybackslash}p{0.12\textwidth}
    }
        \toprule
        \multicolumn{7}{c}{\textbf{Telco Domain Benchmark}} \\
        \midrule
        \textbf{Model}& \textbf{Single QA} & \textbf{Chat CLS} & \textbf{Passage Chat} & \textbf{Chat SM} & \textbf{Vocabulary} & \textbf{Avg.} \\
        \midrule
        \multicolumn{7}{l}{\textbf{Original Instruction Model (Vanilla)}} \\  
        Llama 3.2 3B IT  & 37.00 & 38.00 & 79.33 & 38.50 & 47.00 & 47.97 \\
        Qwen 2.5 3B IT  & 29.00 & 29.50 & 76.67 & 44.71 & 55.50 & 47.08 \\

        \exaone 3.5 2.4B IT  & 16.00 & 58.50 & 42.00 & 40.98 & 54.00 & 42.30 \\ 
        \hdashline
        Llama 3.1 8B IT & 37.00 & 41.50 & 77.33 & 43.39 & 55.50 & 50.90 \\
        Qwen 2.5 7B IT & 25.00 & 67.50 & 62.67 & 36.96 & 56.00 & 49.63 \\
        \exaone 3.5 7.8B IT & 19.00 & 66.00 & 33.33 & 37.27 & 59.00 & 42.92 \\
        \hdashline
        \exaone 3.5 32B IT & 25.00 & 78.00 & 40.67 & 48.43 & 57.00 & 49.82 \\

        \midrule
        \textbf{Telco DACP Model (Ours)} \\
        Llama 3.2  Telco 3B IT&     \textbf{97.00} & \textbf{61.00} & \textbf{80.67} & \textbf{54.75} & \textbf{68.50} & \textbf{72.38 (+51\%)} \\
        Qwen 2.5  Telco 3B IT&      \textbf{94.00} & \textbf{53.00} & \textbf{81.33} & \textbf{63.82} & \textbf{62.50} & \textbf{70.93 (+50\%)} \\
        \exaone 3.5  Telco 2.4B IT& \textbf{96.00} & 56.00 & \textbf{72.00} & \textbf{63.69} & \textbf{67.00} & \textbf{70.94 (+67\%)} \\  
        \hdashline
        Llama 3.2  Telco 8B IT&     \textbf{92.00} & \textbf{72.00} & \textbf{87.33} & \textbf{65.49} & \textbf{69.00} & \textbf{77.16 (+52\%)} \\
        Qwen 2.5  Telco 7B IT&      \textbf{92.00} & 41.50 & \textbf{90.00} & \textbf{59.32} & \textbf{68.50} & \textbf{70.26 (+41\%)} \\  
        \exaone 3.5  Telco 7.8B IT& \textbf{91.00} & 63.50 & \textbf{82.00} & \textbf{53.08} & \textbf{66.00} & \textbf{71.12 (+66\%)} \\
        \hdashline
        \exaone 3.5 Telco 32B IT  & \textbf{94.00} & \textbf{92.00} & \textbf{82.67} & \textbf{60.97} & \textbf{92.50} & \textbf{84.43 (+69\%)} \\
    
        \bottomrule
    \end{tabular}
    \vspace{-1mm}
    \caption{Performance comparison of original instruction models and Telco Domain Adapted models (Telco DACP) evaluated on the Telco Domain Benchmark. Models are grouped by size (dashed lines). Percentage improvements over original models are shown in parentheses. \textbf{Bold} scores indicate an improvement over the instruction model.}
    \vspace{-3mm}
    \label{tab:telcoBench}
\end{table*}

\subsection{Post-training Dataset Preparation}



Since DACP focuses on domain knowledge acquisition rather than instruction-following, it is applied prior to instruction tuning. As a result, DACP-applied models may lack instruction-following ability, necessitating post-training (e.g., instruction and alignment tuning) to enable effective knowledge use. Without this step, models may fail to apply acquired domain knowledge. For post-training, we used public instruction datasets such as Tulu 3 \cite{lambert2025tulu3pushingfrontiers}, AIHub \cite{aihub2025}, and a synthetic dataset generated from human-annotated seed data.


\section{Experiment}

We conducted extensive experiments to validate our method, with training settings in Appendix \ref{sec:training_settings}. This section confirms that DACP applied to Telco and financial datasets consistently improved domain adaptation performance while preserving general domain capabilities.

\subsection{Benchmark Datasets}

The primary objective of DACP is to enhance performance in the target domain while reducing performance degradation in the general domain. 

To verify the maintenance of general domain performance, we utilized publicly available benchmark datasets. Specifically, to evaluate performance in both Korean and English, we used a total of five benchmark datasets: \textit{MMLU}\cite{hendrycks2021measuringmassivemultitasklanguage} for overall language understanding, \textit{KMMLU}\cite{son2024kmmlumeasuringmassivemultitask} and \textit{HAE-RAE} Bench \cite{son2024haeraebenchevaluationkorean} for Korean language comprehension, \textit{Big Bench Hard}\cite{suzgun2022challengingbigbenchtaskschainofthought} for reasoning, and \textit{GSM8k-ko}\cite{kuotientgsm8kko} for assessing Korean mathematical abilities. These general benchmarks were selected to reflect the practical requirements of industrial applications.

On the other hand, for the Telco domain, due to the absence of publicly available benchmarks, we developed an in-house benchmark. The Telco benchmark is primarily based on call center data, and aside from a telecommunications and network terminology test designed to evaluate Telco-related knowledge, it mainly comprises tasks related to call center operations. The detailed composition of the Telco benchmark is presented in Table \ref{tab:TaskExplain}, with examples provided in Appendix \ref{sec:telco_benchmark_details}.

\begin{table}[t]
    \centering
    \footnotesize
    \renewcommand{\arraystretch}{1.1}
    \begin{tabular}{
        >{\raggedright\arraybackslash}p{0.21\textwidth}
        >{\centering\arraybackslash}p{0.10\textwidth}
        >{\centering\arraybackslash}p{0.10\textwidth}
    }
        \toprule
        \textbf{Model} & \textbf{General Avg.} & \textbf{Finance Avg.} \\
        \midrule
        \exaone 3.5 2.4B IT  & 52.64 & 39.52 \\
        \hdashline
        Llama 3.1 8B IT   & 59.43 & 47.40 \\
        Qwen 2.5 7B IT  & 60.74 & 46.98 \\
        \exaone 3.5 7.8B IT  & 60.10 & 45.53 \\
        \hdashline
        Qwen 2.5 32B IT   & 58.48 & 50.48 \\
        \exaone 3.5 32B IT  & 68.15 & 50.70 \\
        \hdashline
        Llama 3.3 70B IT   & 66.05 & 52.23 \\
        Qwen 2.5 72B IT   & 65.37 & 52.62 \\
        \midrule
        \exaone Finance 2.4B IT & 49.64(-6\%) & \textbf{58.88(+49\%)} \\
        \hdashline
        \exaone Finance 7.8B IT  & 59.56(-1\%) & \textbf{59.54(+31\%)} \\
        \bottomrule
    \end{tabular}
    \vspace{-1mm}
    \caption{Performance of instruction models vs. Finance DACP-applied models on Finance and General benchmarks. Bold indicates improvement; percentages show relative gains. DACP-applied to \exaone 3.5 models (2.4B, 7.8B).}
    \vspace{-4mm}
    \label{tab:financeBench}
\end{table}

Unlike the Telco domain, where publicly available benchmarks are scarce, the financial domain benefits from well-established benchmarks due to extensive prior research, some of which were used in our evaluation \cite{islam2023financebench,NEURIPS2024_adb1d9fa,lin2025open}. Detailed information on the financial domain benchmark is provided in Appendix \ref{sec:finance_benchmark_details}.

\subsection{Evaluations}


To validate DACP's effectiveness, we conducted benchmark evaluations to quantitatively assess overall model performance and industrial application evaluations to verify its practical applicability in real-world settings.

\subsubsection{Benchmark Evaluations}




As shown in Figure~\ref{fig:introduction_graph}, applying DACP with Telco domain data to the \exaone 3.5 7.8B foundation model, followed by post-training, resulted in \exaone Telco 7.8B IT, which outperformed both \exaone 3.5 7.8B IT and larger open model competitors on Telco benchmarks. This confirms that domain knowledge was successfully acquired through DACP while maintaining general domain performance comparable to the original \exaone 3.5 7.8B IT model, effectively mitigating catastrophic forgetting.



Building on these findings, Table \ref{tab:generalBench} and \ref{tab:telcoBench} show that applying the same dataset and training methodology through DACP and post-training led to consistent domain performance gains across all models, with only minor fluctuations in general-domain results. Notably, model size moderates the trade-off: very small models (≤3B) can lose up to 7\% on general reasoning benchmarks like BBH, indicating stronger forgetting, whereas larger models both improve domain performance and better preserve—or even enhance—general capabilities. This underscores the need for careful replay and data composition in smaller models and highlights the relative stability of DACP in larger LLMs.


To further assess the generality of our approach beyond a single domain, we additionally applied DACP and post-training to a financial domain dataset. As shown in Table \ref{tab:financeBench}, the proposed approach is effective beyond the Telco domain, with the finance-adapted model outperforming larger general sLLMs in its target domain, similar to the Telco case.




Experiments demonstrate that DACP allows enterprises to deploy domain-adapted sLLMs instead of larger models for cost-efficient operations.

\begin{table*}[t]
    \centering
    \footnotesize
    \renewcommand{\arraystretch}{1.2} 
    \setlength{\tabcolsep}{4.5pt}       
    \begin{tabular}{p{3.9cm} c c c c c c c}
        \toprule
        \textbf{Model} & \textbf{Post} & \textbf{Extraction \&} & \textbf{Computation \&} & \textbf{Strategic} & \textbf{Text} & \textbf{Overall} \\
        & \textbf{Training} & \ \textbf{Summarization} & \textbf{Table Recognition} & \textbf{Planning} & \textbf{Generation} & \textbf{MRR} \\
        \midrule
        \exaone 3.5 7.8B (vanilla) & X & 61.11 & 20.00 & 48.33 & 61.11 & 47.64 \\
        \exaone 3.5 7.8B & O & 34.44 & 50.00 & 44.44 & 21.67 & 37.64 \\
        \rowcolor[gray]{0.9} \exaone 3.5 Finance 7.8B & O & \textbf{83.33} & \textbf{77.78} & \textbf{50.00} & \textbf{83.33} & \textbf{73.61} \\
        \bottomrule
    \end{tabular}
   \vspace{-2mm}
    \caption{We performed an ablation study on each SFT(RAG) and DPO and show the results. Mean Reciprocal Rank (MRR) Evaluation Across Machine Reading Comprehension(MRC) Finance Tasks (rescaled from 0-1 to 0-100 for better clarity) } 
   \vspace{-5mm}

    \label{tab:mrr_evaluation}
\end{table*}

\begin{table}[t]
    \centering
    \scriptsize 
    \setlength{\tabcolsep}{2pt}       
    \begin{tabular}{
        >{\raggedright\arraybackslash}p{0.16\textwidth}
        >{\centering\arraybackslash}p{0.06\textwidth}
        >{\centering\arraybackslash}p{0.06\textwidth}
        >{\centering\arraybackslash}p{0.05\textwidth}
        >{\centering\arraybackslash}p{0.05\textwidth}
        >{\centering\arraybackslash}p{0.03\textwidth}
    }
        \toprule
        & \multicolumn{2}{c}{\textbf{Customer Request}} & \multicolumn{2}{c}{\textbf{Agent Response}} & \\
        \midrule
        \textbf{Model} & \textbf{Mobile} & \textbf{Home} & \textbf{Mobile} & \textbf{Home} & \textbf{Avg.} \\
        \midrule
        \exaone 3.5 2.4B & 4.16 & 4.44 & 3.60 & 4.32 & 4.13 \\
        \rowcolor[gray]{0.9} \exaone 3.5 Telco 2.4B& \textbf{4.52} & \textbf{4.72} & \textbf{4.32} & \textbf{4.44} & \textbf{4.50} \\
        \bottomrule
    \end{tabular}
    \vspace{-2mm}

    \caption{Human evaluation (1–5 scale) of baseline and Telco models on customer service summarization.}
    \vspace{-3mm}

    \label{tab:aicc_eval}
\end{table}



\subsubsection{Evaluations of Telco Application}

\begin{figure}[h]
  \centering
  \vspace{-2mm}
  \includegraphics[width=\linewidth]{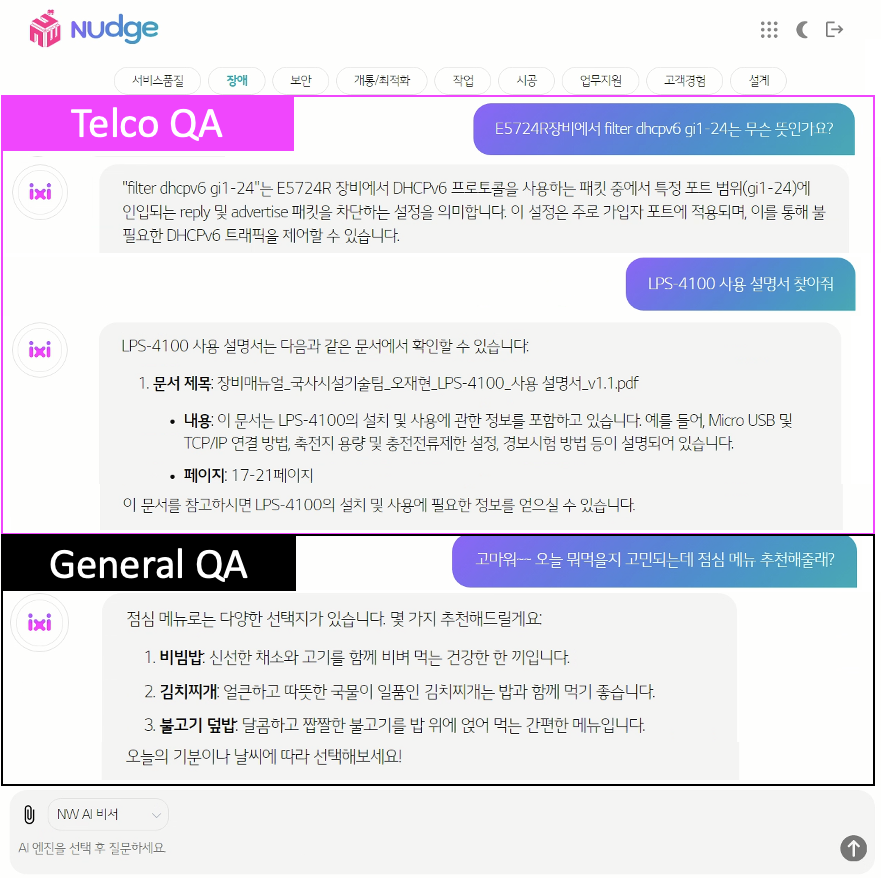}
  \caption{Screenshot and dialogue example of NW QA system. Industrial sLLM usage requires balancing domain adaptation with general knowledge, as overfitting to a domain can harm overall user experience. The translated version is provided in Table~\ref{tab:figure3_conversation_translation} in Appendix.}
  \vspace{-2mm}
  \label{fig:why_we_need_dacp}
\end{figure}

In this study, we evaluated the practical utility of a Telco‑domain LLM enhanced via DACP by deploying it in a real‑world industrial setting—specifically to support customer service agents. After each customer call, agents perform two summarization tasks based on the call transcription: (1) summarizing the customer’s request, and (2) summarizing the actions taken by the agent. We applied Telco adapted-LLM to automate both tasks.


To assess the efficiency of DACP in an industry application, we fine-tuned a baseline foundation model and a Telco-domain adapted model on call summarization tasks, followed by human evaluation (Appendix \ref{sec:industry_application_evaluation_details}). As shown in Table \ref{tab:aicc_eval}, the Telco-domain model significantly outperformed the baseline, demonstrating that exposure to Telco-domain call transcription data enhances the model’s ability to understand and accurately summarize customer service conversations.

\begin{figure} [h]
    \centering
    \includegraphics[width=1\linewidth]{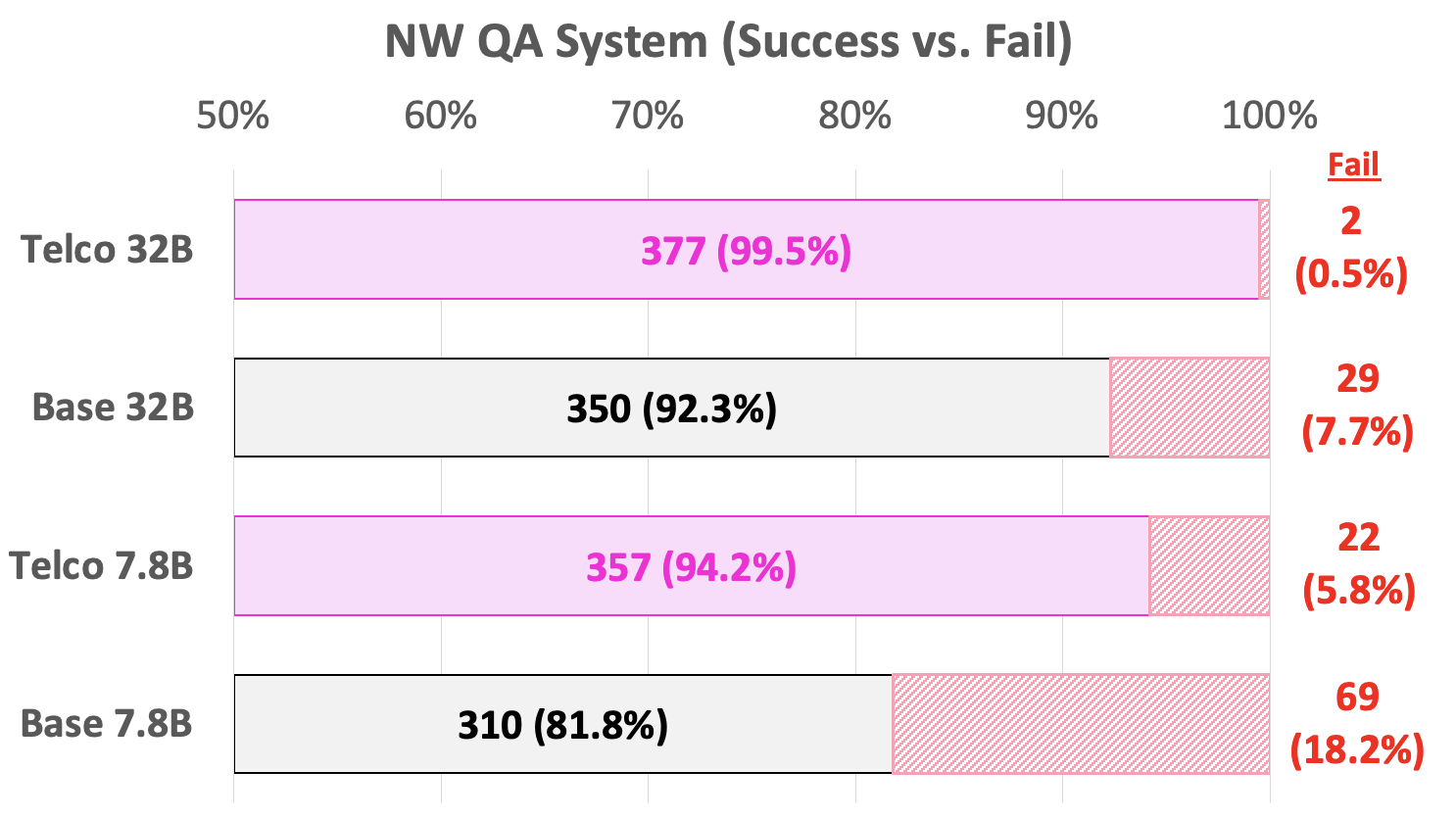}
    \caption{Success / fail rates of telco-adapted vs. base QA models after service SFT, highlighting the gain from Telco adaptation.}
    \vspace{-2mm}
    \label{fig:nw_agent_failure_rates}
\end{figure}

We further applied the Telco-domain adapted model to a network equipment QA system. Both the baseline foundation model and the DACP-applied model were fine-tuned for retrieval-augmented generation (RAG) on the network QA task. As shown in Figure~\ref{fig:nw_agent_failure_rates}, the DACP-applied model showed a substantial reduction in failure rates, demonstrating its effectiveness in enhancing domain-specific understanding. These results confirm that DACP offers a cost-efficient and practical solution for improving RAG performance in industrial settings beyond what is achievable through conventional fine-tuning. 

Moreover, the results show that the smaller DACP-applied model outperforms a larger model, highlighting its practicality for real-world deployments with infrastructure or service-level constraints.


\subsubsection{Evaluation of Finance Application}

We evaluated LLMs with DACP and RAG on financial-domain tasks, as the sector’s specialized terminology and complex concepts provide an effective testbed for assessing DACP. 

As shown in Table \ref{tab:mrr_evaluation}, all models except the first vanilla baseline were post-trained and evaluated using the same top-3 retrieved reference passages. The DACP$+$post-training approach achieved 73.61\% MRR, outperforming both the vanilla \exaone model and the post-trained baseline model, resulting in a 95\% improvement and demonstrating the effectiveness of domain adaptation in specialized financial applications.

\section{Conclusion}

We propose a recipe for executing DACP using a mid-scale domain corpus. Through experiments, we have demonstrated that this methodology can be effectively applied to industrial applications. Our study shows that even in environments with limited inference computing infrastructure, employing a DACP-applied sLLM with a relatively small number of parameters can achieve higher performance than larger parameter models. The proposed approach has proven robust across variations in domain, parameter size, and foundation model type.


This method allows companies to obtain a high-performing, domain-adapted sLLM at a lower cost through DACP, eliminating the need to deploy larger models to meet service quality requirements. By delivering excellent performance in real-world services, it is expected to provide an enhanced user experience to customers.

\section{Limitations}

This study applies DACP to open-source LLMs such as LLaMA, Qwen2.5, and \exaone. However, since most open-source LLMs do not disclose their original pretraining corpora, this poses a structural limitation when constructing replay datasets. To address this, we built a replay corpus for general knowledge retention by combining widely used public sources such as FineWeb, CCNet, and Wikipedia. While this approach is practical in terms of reproducibility and accessibility, the resulting data may differ from the original training data in topic coverage, writing style, and informational depth.

In particular, if the replay corpus is not sufficiently aligned with the general knowledge on which the base model was originally trained, the DACP process may still nominally perform replay, but may fail to retain general knowledge effectively—thereby undermining the intended effect of knowledge preservation. To mitigate this, future work should focus on indirectly estimating the characteristics of the original pretraining data and designing more precisely aligned replay corpora that better match its distribution and style.

\bibliography{custom}

\begin{thebibliography}{49}
\providecommand{\natexlab}[1]{#1}

\bibitem[{{AI Hub}(2025)}]{aihub2025}
{AI Hub}. 2025.
\newblock Ai hub: Artificial intelligence data and services portal.
\newblock \url{https://www.aihub.or.kr}.
\newblock Accessed: 2025-07-04.

\bibitem[{Allganize(2024)}]{allganize2024rag}
Allganize. 2024.
\newblock Rag evaluation dataset ko.
\newblock
  \url{https://huggingface.co/datasets/allganize/RAG-Evaluation-Dataset-KO}.
\newblock Accessed: 2025-03-21.

\bibitem[{{Common Crawl}(2025)}]{commoncrawl2025}
{Common Crawl}. 2025.
\newblock Common crawl: Open web crawling data.
\newblock \url{https://commoncrawl.org}.
\newblock Accessed: 2025-07-04.

\bibitem[{Cui et~al.(2024)Cui, Yang, and
  Yao}]{cui2024efficienteffectivetextencoding}
Yiming Cui, Ziqing Yang, and Xin Yao. 2024.
\newblock \href {https://arxiv.org/abs/2304.08177} {Efficient and effective
  text encoding for chinese llama and alpaca}.
\newblock \emph{Preprint}, arXiv:2304.08177.

\bibitem[{Gao et~al.(2021)Gao, Biderman, Black, Golding, Heimbigner
  et~al.}]{evalharness2021}
Leo Gao, Stella Biderman, Sid Black, Charles Golding, Eric Heimbigner, et~al.
  2021.
\newblock lm-evaluation-harness.
\newblock \url{https://github.com/EleutherAI/lm-evaluation-harness}.
\newblock Accessed: 2025-03-21.

\bibitem[{Gekhman et~al.(2024)Gekhman, Yona, Aharoni, Eyal, Feder, Reichart,
  and Herzig}]{gekhman-etal-2024-does}
Zorik Gekhman, Gal Yona, Roee Aharoni, Matan Eyal, Amir Feder, Roi Reichart,
  and Jonathan Herzig. 2024.
\newblock \href {https://aclanthology.org/2024.emnlp-main.444} {Does
  fine-tuning llms on new knowledge encourage hallucinations?}
\newblock In \emph{Proceedings of the 2024 Conference on Empirical Methods in
  Natural Language Processing}, pages 7765--7784, Singapore. Association for
  Computational Linguistics.

\bibitem[{{Gemma Team}(2024)}]{gemma2025gemma3}
Google~DeepMind {Gemma Team}. 2024.
\newblock \href
  {https://storage.googleapis.com/deepmind-media/gemma/Gemma3Report.pdf}
  {Gemma: Open models based on gemini research and technology}.
\newblock Technical report, Google DeepMind.
\newblock Accessed: 2025-03-21.

\bibitem[{Grattafiori et~al.(2024)Grattafiori, Dubey, Jauhri, Pandey, Kadian,
  Al-Dahle, Letman, Mathur, Schelten, Vaughan
  et~al.}]{grattafiori2024llama3herdmodels}
Aaron Grattafiori, Abhimanyu Dubey, Abhinav Jauhri, Abhinav Pandey, Abhishek
  Kadian, Ahmad Al-Dahle, Aiesha Letman, Akhil Mathur, Alan Schelten, Alex
  Vaughan, et~al. 2024.
\newblock \href {https://arxiv.org/abs/2407.21783} {The llama 3 herd of
  models}.

\bibitem[{Guo et~al.(2025)Guo, Yang, Zhang, Song, Zhang, Xu, Zhu, Ma, Wang, Bi
  et~al.}]{deepseekai2025deepseekr1incentivizingreasoningcapability}
Daya Guo, Dejian Yang, Haowei Zhang, Junxiao Song, Ruoyu Zhang, Runxin Xu,
  Qihao Zhu, Shirong Ma, Peiyi Wang, Xiao Bi, et~al. 2025.
\newblock \href {https://arxiv.org/abs/2501.12948} {Deepseek-r1: Incentivizing
  reasoning capability in llms via reinforcement learning}.
\newblock \emph{Preprint}, arXiv:2501.12948.

\bibitem[{Gupta et~al.(2023)Gupta, Thérien, Ibrahim, Richter, Anthony,
  Belilovsky, Rish, and Lesort}]{gupta2023continualpretraininglargelanguage}
Kshitij Gupta, Benjamin Thérien, Adam Ibrahim, Mats~L. Richter, Quentin
  Anthony, Eugene Belilovsky, Irina Rish, and Timothée Lesort. 2023.
\newblock \href {https://doi.org/10.48550/arXiv.2308.04014} {Continual
  pre-training of large language models: How to (re)warm your model?}
\newblock \emph{CoRR}, abs/2308.04014.

\bibitem[{Gururangan et~al.(2020)Gururangan, Marasovi{\'c}, Swayamdipta, Lo,
  Beltagy, Downey, and Smith}]{gururangan2020dontstoppretrainingadapt}
Suchin Gururangan, Ana Marasovi{\'c}, Swabha Swayamdipta, Kyle Lo, Iz~Beltagy,
  Doug Downey, and Noah~A Smith. 2020.
\newblock \href {https://aclanthology.org/2020.acl-main.740} {Don’t stop
  pretraining: Adapt language models to domains and tasks}.
\newblock In \emph{Proceedings of the 58th Annual Meeting of the Association
  for Computational Linguistics}, pages 8342--8360.

\bibitem[{Hendrycks et~al.(2021)Hendrycks, Burns, Basart, Zou, Mazeika, Song,
  and Steinhardt}]{hendrycks2021measuringmassivemultitasklanguage}
Dan Hendrycks, Collin Burns, Steven Basart, Andy Zou, Mantas Mazeika, Dawn
  Song, and Jacob Steinhardt. 2021.
\newblock \href {https://openreview.net/forum?id=d7KBjmI3GmQ} {Measuring
  massive multitask language understanding}.
\newblock In \emph{International Conference on Learning Representations}.

\bibitem[{{Hugging Face}(2024)}]{fineweb2024}
{Hugging Face}. 2024.
\newblock Fineweb: A high-quality web-scale english text corpus.
\newblock \url{https://huggingface.co/datasets/HuggingFaceFW/fineweb}.
\newblock Accessed: 2025-07-04.

\bibitem[{Islam et~al.(2023)Islam, Kannappan, Kiela, Qian, Scherrer, and
  Vidgen}]{islam2023financebench}
Pranab Islam, Anand Kannappan, Douwe Kiela, Rebecca Qian, Nino Scherrer, and
  Bertie Vidgen. 2023.
\newblock \href {https://arxiv.org/abs/2311.11944} {Financebench: A new
  benchmark for financial question answering}.
\newblock \emph{Preprint}, arXiv:2311.11944.

\bibitem[{Jang et~al.(2022)Jang, Ye, Lee, Yang, Shin, Han, Kim, and
  Seo}]{jang2023temporalwikilifelongbenchmarktraining}
Joel Jang, Seonghyeon Ye, Changho Lee, Sohee Yang, Joongbo Shin, Janghoon Han,
  Gyeonghun Kim, and Minjoon Seo. 2022.
\newblock \href {https://aclanthology.org/2022.emnlp-main.418} {Temporalwiki: A
  lifelong benchmark for training and evaluating ever-evolving language
  models}.
\newblock In \emph{Proceedings of the 2022 Conference on Empirical Methods in
  Natural Language Processing}, pages 6237--6250.

\bibitem[{Jiang et~al.(2024)Jiang, Huang, Luo, Zhang, Huang, Wei, Deng, Sun,
  Zhang, Wang et~al.}]{jiang2024improving}
Ting Jiang, Shaohan Huang, Shengyue Luo, Zihan Zhang, Haizhen Huang, Furu Wei,
  Weiwei Deng, Feng Sun, Qi~Zhang, Deqing Wang, et~al. 2024.
\newblock Improving domain adaptation through extended-text reading
  comprehension.
\newblock \emph{arXiv preprint arXiv:2401.07284}.

\bibitem[{Kaur et~al.(2024)Kaur, Kashyap, Kumar, Nafis, Kumar, and
  Shokeen}]{kaur2024texttransformationcomprehensivereview}
Pravneet Kaur, Gautam~Siddharth Kashyap, Ankit Kumar, Md~Tabrez Nafis, Sandeep
  Kumar, and Vikrant Shokeen. 2024.
\newblock \href {https://arxiv.org/abs/2402.16142} {From text to
  transformation: A comprehensive review of large language models'
  versatility}.
\newblock \emph{Preprint}, arXiv:2402.16142.

\bibitem[{Kim et~al.(2022)Kim, Jang, Kwon, and
  Davis}]{kim2022kobestkoreanbalancedevaluation}
Dohyeong Kim, Myeongjun Jang, Deuk~Sin Kwon, and Eric Davis. 2022.
\newblock \href {https://arxiv.org/abs/2204.04541} {Kobest: Korean balanced
  evaluation of significant tasks}.
\newblock \emph{Preprint}, arXiv:2204.04541.

\bibitem[{Kim et~al.(2024)Kim, Choi, and
  Jeong}]{kim2024efficienteffectivevocabularyexpansion}
Seungduk Kim, Seungtaek Choi, and Myeongho Jeong. 2024.
\newblock \href {https://arxiv.org/abs/2402.14714} {Efficient and effective
  vocabulary expansion towards multilingual large language models}.
\newblock \emph{Preprint}, arXiv:2402.14714.

\bibitem[{Kuotient(2024)}]{kuotientgsm8kko}
Kuotient. 2024.
\newblock \href {https://huggingface.co/datasets/kuotient/gsm8k-ko} {Gsm8k-ko:
  A korean version of the gsm8k dataset}.
\newblock Accessed: 2025-03-21.

\bibitem[{Kwon et~al.(2023)Kwon, Li, Zhuang, Sheng, Zheng, Yu, Gonzalez, Zhang,
  and Stoica}]{kwon2023efficient}
Woosuk Kwon, Zhuohan Li, Siyuan Zhuang, Ying Sheng, Lianmin Zheng, Cody~Hao Yu,
  Joseph Gonzalez, Hao Zhang, and Ion Stoica. 2023.
\newblock \href {https://dl.acm.org/doi/abs/10.1145/3600006.3613165} {Efficient
  memory management for large language model serving with pagedattention}.
\newblock In \emph{Proceedings of the 29th Symposium on Operating Systems
  Principles}, pages 611--626.

\bibitem[{Lambert et~al.(2025)Lambert, Morrison, Pyatkin, Huang, Ivison,
  Brahman, Miranda, Liu, Dziri, Lyu, Gu, Malik, Graf, Hwang, Yang, Bras,
  Tafjord, Wilhelm, Soldaini, Smith, Wang, Dasigi, and
  Hajishirzi}]{lambert2025tulu3pushingfrontiers}
Nathan Lambert, Jacob Morrison, Valentina Pyatkin, Shengyi Huang, Hamish
  Ivison, Faeze Brahman, Lester James~V. Miranda, Alisa Liu, Nouha Dziri, Shane
  Lyu, Yuling Gu, Saumya Malik, Victoria Graf, Jena~D. Hwang, Jiangjiang Yang,
  Ronan~Le Bras, Oyvind Tafjord, Chris Wilhelm, Luca Soldaini, Noah~A. Smith,
  Yizhong Wang, Pradeep Dasigi, and Hannaneh Hajishirzi. 2025.
\newblock \href {https://arxiv.org/abs/2411.15124} {Tulu 3: Pushing frontiers
  in open language model post-training}.
\newblock \emph{Preprint}, arXiv:2411.15124.

\bibitem[{Li et~al.(2024)Li, Zhang, Bubeck, Pathuri, and
  Menache}]{li2024smalllanguagemodelsapplication}
Beibin Li, Yi~Zhang, Sébastien Bubeck, Jeevan Pathuri, and Ishai Menache.
  2024.
\newblock \href {https://arxiv.org/abs/2405.20347} {Small language models for
  application interactions: A case study}.
\newblock \emph{Preprint}, arXiv:2405.20347.

\bibitem[{Li and Hoiem(2018)}]{li2017learningforgetting}
Zhizhong Li and Derek Hoiem. 2018.
\newblock \href {https://doi.org/10.1109/TPAMI.2017.2773081} {Learning without
  forgetting}.
\newblock \emph{IEEE Transactions on Pattern Analysis and Machine
  Intelligence}, 40(12):2935--2947.

\bibitem[{Liao et~al.(2024)Liao, Xie, Sun, Sun, and
  Kang}]{liao2024exploringforgettinglargelanguage}
Chonghua Liao, Ruobing Xie, Xingwu Sun, Haowen Sun, and Zhanhui Kang. 2024.
\newblock \href {https://arxiv.org/abs/2410.17018} {Exploring forgetting in
  large language model pre-training}.
\newblock \emph{Preprint}, arXiv:2410.17018.

\bibitem[{Lin et~al.(2025)Lin, Tian, Wang, Zhao, Huang, Xie, Borella, White,
  Wang, Xiao et~al.}]{lin2025open}
Shengyuan~Colin Lin, Felix Tian, Keyi Wang, Xingjian Zhao, Jimin Huang,
  Qianqian Xie, Luca Borella, Matt White, Christina~Dan Wang, Kairong Xiao,
  et~al. 2025.
\newblock \href {https://arxiv.org/abs/2501.10963} {Open finllm leaderboard:
  Towards financial ai readiness}.
\newblock \emph{Preprint}, arXiv:2501.10963.

\bibitem[{Ling et~al.(2023)Ling, Zhao, Lu, Deng, Zheng, Wang, Chowdhury, Li,
  Cui, Zhang et~al.}]{ling2024domainspecializationkeymake}
Chen Ling, Xujiang Zhao, Jiaying Lu, Chengyuan Deng, Can Zheng, Junxiang Wang,
  Tanmoy Chowdhury, Yun Li, Hejie Cui, Xuchao Zhang, et~al. 2023.
\newblock \href {https://arxiv.org/abs/2305.18703} {Domain specialization as
  the key to make large language models disruptive: A comprehensive survey}.
\newblock \emph{Preprint}, arXiv:2305.18703.

\bibitem[{Liu et~al.(2024)Liu, Kang, Wang, Qing, Zhao, Wu, Sun, Kuang, and
  Wu}]{liu-etal-2024-catastrophic}
Chengyuan Liu, Yangyang Kang, Shihang Wang, Lizhi Qing, Fubang Zhao, Chao Wu,
  Changlong Sun, Kun Kuang, and Fei Wu. 2024.
\newblock \href {https://doi.org/10.18653/v1/2024.emnlp-main.429} {More than
  catastrophic forgetting: Integrating general capabilities for domain-specific
  {LLM}s}.
\newblock In \emph{Proceedings of the 2024 Conference on Empirical Methods in
  Natural Language Processing}, pages 7531--7548.

\bibitem[{Lu et~al.(2024{\natexlab{a}})Lu, Luu, and
  Buehler}]{lu2024finetuninglargelanguagemodels}
Wei Lu, Rachel~K. Luu, and Markus~J. Buehler. 2024{\natexlab{a}}.
\newblock \href {https://arxiv.org/abs/2409.03444} {Fine-tuning large language
  models for domain adaptation: Exploration of training strategies, scaling,
  model merging and synergistic capabilities}.
\newblock \emph{Preprint}, arXiv:2409.03444.

\bibitem[{Lu et~al.(2024{\natexlab{b}})Lu, Li, Cai, Yi, Liu, Zhang, Lane, and
  Xu}]{lu2025smalllanguagemodelssurvey}
Zhenyan Lu, Xiang Li, Dongqi Cai, Rongjie Yi, Fangming Liu, Xiwen Zhang,
  Nicholas~D. Lane, and Mengwei Xu. 2024{\natexlab{b}}.
\newblock \href {https://doi.org/10.48550/arXiv.2409.15790} {Small language
  models: Survey, measurements, and insights}.
\newblock \emph{CoRR}, abs/2409.15790.

\bibitem[{{National Institute of the Korean Language}(2025)}]{klri2025}
{National Institute of the Korean Language}. 2025.
\newblock Korean language resource portal.
\newblock \url{https://www.korean.go.kr}.
\newblock Accessed: 2025-07-04.

\bibitem[{Parisi et~al.(2019)Parisi, Kemker, Part, Kanan, and
  Wermter}]{parisi2019continual}
German~I. Parisi, Ronald Kemker, Jose~L. Part, Christopher Kanan, and Stefan
  Wermter. 2019.
\newblock \href {https://doi.org/10.1016/j.neunet.2019.01.012} {Continual
  lifelong learning with neural networks: A review}.
\newblock \emph{Neural Networks}, 113:54--71.

\bibitem[{Sanh et~al.(2022)Sanh, Webson, Raffel, Bach, Sutawika, Alyafeai,
  Chaffin, Stiegler, Raja, Dey, Bari et~al.}]{sanh2021multitask}
Victor Sanh, Albert Webson, Colin Raffel, Stephen Bach, Lintang Sutawika, Zaid
  Alyafeai, Antoine Chaffin, Arnaud Stiegler, Arun Raja, Manan Dey, M~Saiful
  Bari, et~al. 2022.
\newblock \href {https://openreview.net/forum?id=9Vrb9D0WI4} {Multitask
  prompted training enables zero-shot task generalization}.
\newblock In \emph{International Conference on Learning Representations}.

\bibitem[{Scialom et~al.(2022)Scialom, Chakrabarty, and
  Muresan}]{scialom2022finetunedlanguagemodelscontinual}
Thomas Scialom, Tuhin Chakrabarty, and Smaranda Muresan. 2022.
\newblock \href {https://doi.org/10.18653/v1/2022.emnlp-main.410} {Fine-tuned
  language models are continual learners}.
\newblock In \emph{Proceedings of the 2022 Conference on Empirical Methods in
  Natural Language Processing}, pages 6107--6122, Abu Dhabi, United Arab
  Emirates. Association for Computational Linguistics.

\bibitem[{Singhal et~al.(2023)Singhal, Azizi, Tu, Mahdavi, Wei, Chung, Scales,
  Tanwani, Cole-Lewis, Pfohl et~al.}]{singhal2022largelanguagemodelsencode}
Karan Singhal, Shekoofeh Azizi, Tao Tu, S~Sara Mahdavi, Jason Wei, Hyung~Won
  Chung, Nathan Scales, Ajay Tanwani, Heather Cole-Lewis, Stephen Pfohl, et~al.
  2023.
\newblock Publisher correction: Large language models encode clinical
  knowledge.
\newblock \emph{Nature}, 620(7973):E19--E19.

\bibitem[{Son et~al.(2024{\natexlab{a}})Son, Lee, Kim, Kim, Muennighoff, Choi,
  Park, Yoo, and Biderman}]{son2024kmmlumeasuringmassivemultitask}
Guijin Son, Hanwool Lee, Sungdong Kim, Seungone Kim, Niklas Muennighoff,
  Taekyoon Choi, Cheonbok Park, Kang~Min Yoo, and Stella Biderman.
  2024{\natexlab{a}}.
\newblock \href {https://doi.org/10.48550/arXiv.2402.11548} {Kmmlu: Measuring
  massive multitask language understanding in korean}.
\newblock \emph{CoRR}, abs/2402.11548.

\bibitem[{Son et~al.(2024{\natexlab{b}})Son, Lee, Kim, Kim, cheol Lee, Yeom,
  Jung, woo Kim, and Kim}]{son2024haeraebenchevaluationkorean}
Guijin Son, Hanwool Lee, Suwan Kim, Huiseo Kim, Jae cheol Lee, Je~Won Yeom,
  Jihyu Jung, Jung woo Kim, and Songseong Kim. 2024{\natexlab{b}}.
\newblock \href {https://aclanthology.org/2024.lrec-main.704} {Hae-rae bench:
  Evaluation of korean knowledge in language models}.
\newblock In \emph{Proceedings of the 2024 Joint International Conference on
  Computational Linguistics, Language Resources and Evaluation (LREC-COLING
  2024)}, pages 7993--8007.

\bibitem[{Song et~al.(2025)Song, Yan, Liu, Fang, Li, Yan, and
  Chen}]{song2025injectingdomainspecificknowledgelarge}
Zirui Song, Bin Yan, Yuhan Liu, Miao Fang, Mingzhe Li, Rui Yan, and Xiuying
  Chen. 2025.
\newblock \href {https://arxiv.org/abs/2502.10708} {Injecting domain-specific
  knowledge into large language models: A comprehensive survey}.
\newblock \emph{Preprint}, arXiv:2502.10708.

\bibitem[{Suzgun et~al.(2023)Suzgun, Scales, Sch{\"a}rli, Gehrmann, Tay, Chung,
  Chowdhery, Le, Chi, Zhou
  et~al.}]{suzgun2022challengingbigbenchtaskschainofthought}
Mirac Suzgun, Nathan Scales, Nathanael Sch{\"a}rli, Sebastian Gehrmann, Yi~Tay,
  Hyung~Won Chung, Aakanksha Chowdhery, Quoc Le, Ed~Chi, Denny Zhou, et~al.
  2023.
\newblock \href {https://aclanthology.org/2023.findings-acl.824/} {Challenging
  big-bench tasks and whether chain-of-thought can solve them}.
\newblock In \emph{Findings of the Association for Computational Linguistics:
  ACL 2023}, pages 13003--13051.

\bibitem[{Touvron et~al.(2023)Touvron, Lavril, Izacard, Martinet, Lachaux,
  Lacroix, Rozière, Goyal, Hambro, Azhar, Rodriguez, Joulin, Grave, and
  Lample}]{touvron2023llamaopenefficientfoundation}
Hugo Touvron, Thibaut Lavril, Gautier Izacard, Xavier Martinet, Marie-Anne
  Lachaux, Timothée Lacroix, Baptiste Rozière, Naman Goyal, Eric Hambro,
  Faisal Azhar, Aurelien Rodriguez, Armand Joulin, Edouard Grave, and Guillaume
  Lample. 2023.
\newblock \href {https://arxiv.org/abs/2302.13971} {Llama: Open and efficient
  foundation language models}.
\newblock \emph{Preprint}, arXiv:2302.13971.

\bibitem[{van~de Ven and Tolias(2019)}]{vandeven2019scenarioscontinuallearning}
Gido~M. van~de Ven and Andreas~S. Tolias. 2019.
\newblock \href {https://arxiv.org/abs/1904.07734} {Three scenarios for
  continual learning}.
\newblock \emph{Preprint}, arXiv:1904.07734.

\bibitem[{Wu et~al.(2023)Wu, Irsoy, Lu, Dabravolski, Dredze, Gehrmann,
  Kambadur, Rosenberg, and Mann}]{wu2023bloomberggpt}
Shijie Wu, Ozan Irsoy, Steven Lu, Vadim Dabravolski, Mark Dredze, Sebastian
  Gehrmann, Prabhanjan Kambadur, David Rosenberg, and Gideon Mann. 2023.
\newblock \href {https://arxiv.org/abs/2303.17564} {Bloomberggpt: A large
  language model for finance}.
\newblock \emph{Preprint}, arXiv:2303.17564.

\bibitem[{Wu et~al.(2024)Wu, Luo, Li, Pan, Vu, and
  Haffari}]{wu2024continuallearninglargelanguage}
Tongtong Wu, Linhao Luo, Yuan-Fang Li, Shirui Pan, Thuy-Trang Vu, and
  Gholamreza Haffari. 2024.
\newblock \href {https://doi.org/10.48550/arXiv.2402.01364} {Continual learning
  for large language models: A survey}.
\newblock \emph{CoRR}, abs/2402.01364.

\bibitem[{Xie et~al.(2024{\natexlab{a}})Xie, Han, Chen, Xiang, Zhang, He, Xiao,
  Li, Dai, Feng et~al.}]{NEURIPS2024_adb1d9fa}
Qianqian Xie, Weiguang Han, Zhengyu Chen, Ruoyu Xiang, Xiao Zhang, Yueru He,
  Mengxi Xiao, Dong Li, Yongfu Dai, Duanyu Feng, et~al. 2024{\natexlab{a}}.
\newblock \href
  {https://proceedings.neurips.cc/paper_files/paper/2024/file/adb1d9fa8be4576d28703b396b82ba1b-Paper-Datasets_and_Benchmarks_Track.pdf}
  {Finben: A holistic financial benchmark for large language models}.
\newblock In \emph{Advances in Neural Information Processing Systems},
  volume~37, pages 95716--95743. Curran Associates, Inc.

\bibitem[{Xie et~al.(2024{\natexlab{b}})Xie, Aggarwal, and
  Ahmad}]{xie2023efficientcontinualpretrainingbuilding}
Yong Xie, Karan Aggarwal, and Aitzaz Ahmad. 2024{\natexlab{b}}.
\newblock \href {https://aclanthology.org/2024.findings-acl.606} {Efficient
  continual pre-training for building domain specific large language models}.
\newblock In \emph{Findings of the Association for Computational Linguistics
  ACL 2024}, pages 10184--10201.

\bibitem[{Yang et~al.(2023)Yang, Liu, and
  Wang}]{yang2023fingptopensourcefinanciallarge}
Hongyang Yang, Xiao-Yang Liu, and Christina~Dan Wang. 2023.
\newblock \href {https://arxiv.org/abs/2306.06031} {Fingpt: Open-source
  financial large language models}.
\newblock \emph{Preprint}, arXiv:2306.06031.

\bibitem[{Yogatama et~al.(2019)Yogatama, d'Autume, Connor, Kocisky,
  Chrzanowski, Kong, Lazaridou, Ling, Yu, Dyer et~al.}]{yogatama2019learning}
Dani Yogatama, Cyprien de~Masson d'Autume, Jerome Connor, Tomas Kocisky, Mike
  Chrzanowski, Lingpeng Kong, Angeliki Lazaridou, Wang Ling, Lei Yu, Chris
  Dyer, et~al. 2019.
\newblock \href {https://arxiv.org/abs/1901.11373} {Learning and evaluating
  general linguistic intelligence}.
\newblock \emph{Preprint}, arXiv:1901.11373.

\bibitem[{Zhao et~al.(2024{\natexlab{a}})Zhao, Zhang, Gao, Zhang, Gui, and
  Huang}]{zhao2024llamaenglishempiricalstudy}
Jun Zhao, Zhihao Zhang, Luhui Gao, Qi~Zhang, Tao Gui, and Xuanjing Huang.
  2024{\natexlab{a}}.
\newblock \href {https://arxiv.org/abs/2401.01055} {Llama beyond english: An
  empirical study on language capability transfer}.
\newblock \emph{Preprint}, arXiv:2401.01055.

\bibitem[{Zhao et~al.(2024{\natexlab{b}})Zhao, Zhang, Chen, Kawaguchi, and
  Bing}]{zhao2024largelanguagemodelshandle}
Yiran Zhao, Wenxuan Zhang, Guizhen Chen, Kenji Kawaguchi, and Lidong Bing.
  2024{\natexlab{b}}.
\newblock \href {https://doi.org/10.48550/arXiv.2402.18815} {How do large
  language models handle multilingualism?}
\newblock \emph{CoRR}, abs/2402.18815.

\end{thebibliography}

\appendix

\section{Experimental Details for Replay Ratio Selection in DACP}
\label{sec:replay_ratio_selection_detail}

This experiment was designed to identify the optimal proportion of replay data for use in DACP. The goal was to determine an effective balance between retaining general-domain knowledge and specializing in a target domain, specifically telecommunications. To conduct this experiment efficiently, we used a smaller-scale model, \exaone-3.5 2.4B, and constructed a lightweight dataset of approximately 10GB (around 3 billion tokens) sampled from the full pretraining corpus. Within this fixed data budget, we incrementally increased the proportion of replay data creating five different data mixtures. Each mixture was then used to train the model following a consistent configuration that mirrored the setup of the full DACP process, including similar optimization schedules and training steps.

After training, all five models were evaluated on general-domain benchmarks and five telecommunications-specific benchmarks (see Appendix \ref{sec:telco_benchmark_details} for details). For the general benchmarks, we used non-instruction-following tasks only while all benchmarks of Appendix \ref{sec:general_benchmark_details} and KoBEST\cite{kim2022kobestkoreanbalancedevaluation}, in line with the base model’s capabilities, to focus on raw knowledge retention. In addition, the original, untrained \exaone-3.5 2.4B model was evaluated on the same benchmarks to serve as a baseline for comparison.

The results showed that increasing the replay data ratio led to improvements in general-domain performance, while domain-specific performance gradually declined. In particular, performance gains on the general benchmarks began to saturate once the replay data ratio exceeded 50\%, whereas the degradation in domain-specific performance accelerated beyond this point. Based on this trend, we determined that a 50\% ratio of replay to domain-specific data provided the most balanced outcome—effectively mitigating catastrophic forgetting while still achieving noticeable improvements in domain specialization. This result is illustrated in Figure~\ref{fig:telco_dacp_ratio}, and detailed benchmark results for this experiment are provided in Tables~\ref{tab:ratio_test_general_bench} and~\ref{tab:ratio_test_telco_bench}. Based on the results of this experiment, we selected the 50\% ratio when constructing the full-scale DACP dataset. 

\begin{table*}[t]
    \small
    \centering
    \begin{tabular}{c|cccc|c}

\hline
\toprule

\textbf{Replay Ratio} & \textbf{KMMLU} & \textbf{HAERAE} & \textbf{KoBest} & \textbf{MMLU} & \textbf{General AVG.} \\
\midrule
5\%  & 26.59 & 63.79 & 59.84 & 50.34 & 50.14 \\ 
10\% & 27.10 & 65.44 & 59.97 & 50.22 & 50.68 \\ 
25\% & 27.70 & 64.16 & 61.46 & 49.72 & 50.76 \\
\rowcolor[gray]{0.9}  50\% & 33.00 & 64.53 & 66.29 & 51.37 & 53.80 \\
75\% & 32.73 & 65.35 & 66.87 & 51.32 & 54.07 \\
\midrule
Base & 32.20 & 64.62 & 67.17 & 53.78 & 54.44 \\ 
    \bottomrule
    \end{tabular}
    \caption{General benchmarks performance comparison of replay ratio}
   \label{tab:ratio_test_general_bench}
\end{table*}

\begin{table*}[t]
    \small
    \centering
    \begin{tabular}{c|ccccc|c}
    \hline
    \toprule

\textbf{Replay Ratio} & \textbf{Vocabulary} & \textbf{Single QA} & \textbf{Chat CLS} & \textbf{Chat SM} & \textbf{Passage Chat} & \textbf{Telco AVG.} \\ 
\midrule
5\%  & 59.50 & 66.00 & 31.50 & 63.57 & 84.00 & 60.91 \\ 
10\% & 59.50 & 70.00 & 31.50 & 62.28 & 79.33 & 60.52 \\ 
25\% & 61.00 & 66.00 & 30.00 & 62.95 & 79.33 & 59.86 \\
\rowcolor[gray]{0.9}  50\% & 61.00 & 69.00 & 7.50  & 64.77 & 76.00 & 55.65 \\
75\% & 54.50 & 66.00 & 8.00  & 36.19 & 78.67 & 48.67 \\
\midrule
Base & 56.50 & 39.00 & 14.50 & 44.03 & 67.33 & 44.27 \\ 
    \bottomrule
    \end{tabular}
    \caption{Telco benchmarks performance comparison of replay ratio}
   \label{tab:ratio_test_telco_bench}
\end{table*}

\begin{figure}[h]
  \centering
  \includegraphics[width=\linewidth]{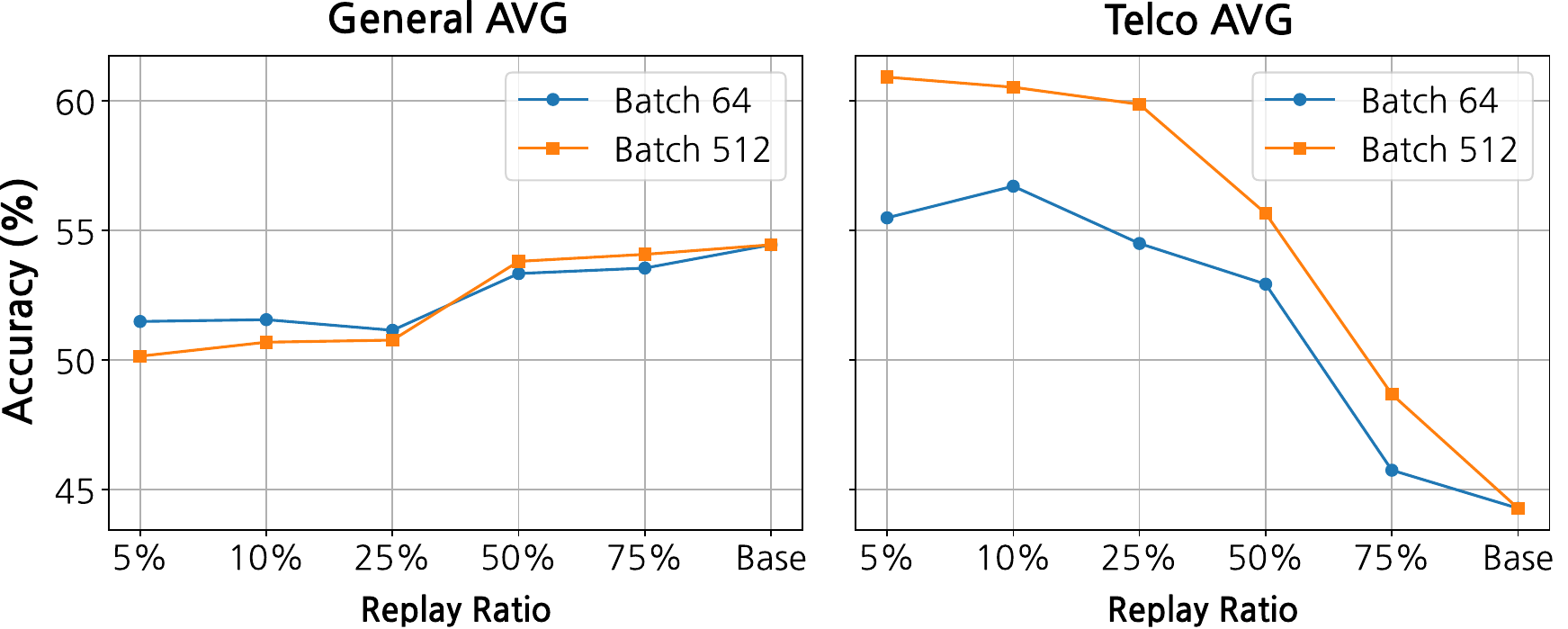}
  \caption{Impact of Replay Ratio and Batch Size on Accuracy.
Average accuracy on General (left) and Telco (right) benchmarks for batch sizes 64 and 512. While accuracy in the General domain increases with higher replay ratios, the Telco domain shows a performance drop beyond 25\%, highlighting differing sensitivities to replay strategies.}
  \label{fig:batch_replay_ratio}
\end{figure}

During the experiment, we applied a reduced training configuration with smaller batch sizes and shorter context lengths to facilitate faster iteration. While this setting slightly limited the degree of domain adaptation and made forgetting effects appear less severe, it proved sufficient for identifying replay ratio trends. Moreover, this setup helped us make informed decisions about learning rate, token budget, and batch strategy for full-scale training. This result is illustrated in Figure~\ref{fig:batch_replay_ratio}. Detailed hyperparameter configurations used in this experiment are documented in Appendix \ref{sec:training_settings}.

\section{Preprocessing in Telco Customer Service Conversation Transcriptions}
\label{sec:telco_preprocessing}

In the case of telecommunications inquiries of DACP datasets, speech-to-text (STT) transcripts of customer service calls were utilized. However, these transcripts inherently contain various STT errors (e.g., typographical errors, recognition failures). To address this, we employed an SFT-trained sLLM, leveraging manually corrected preprocessing data and typo correction datasets. For error correction, this enabled us to reconstruct sanitized data, and the preprocessing process was finalized by performing random sampling-based validation.

Additionally, since the transcripts often contain sensitive personal information, de-identification was necessary. To this end, we developed a dataset for detecting 19 categories of personal information (e.g., names, phone numbers, addresses, dates of birth) in telecommunication transcripts. A BERT-based personal information detection model was trained using this dataset. The detected personal information was subsequently masked through a post-processing logic, and sanitized datasets were reconstructed accordingly. As with the previous step, random sampling-based validation was conducted to complete the preprocessing.

\section{Training Settings}
\label{sec:training_settings}

During continual pretraining, hyperparameters such as the learning rate and the proportion of replay data are critical. An excessively high learning rate risks catastrophic forgetting of pretrained knowledge, whereas an overly low learning rate impedes the effective acquisition of domain knowledge. Similarly, a low replay ratio increases forgetting, while an excessively high ratio reduces domain learning efficiency. We used empirical experimentation to determine a learning rate schedule using a cosine decay scheduler with an initial rate of $1\times10^{-5}$. To mitigate forgetting, we limited the number of training steps by setting the maximum context length to 32K tokens and configured a global batch size of 2,048.

\section{General Benchmark Details}
\label{sec:general_benchmark_details}

We evaluate five general benchmarks (\textit{MMLU, BBH, KMMLU, HARRAE, GSM8K-Ko}) using \textit{LM Eval Harness}\cite{evalharness2021}. Given that generation speed is crucial for enterprise services, we employ the \textit{vLLM} serving framework \cite{kwon2023efficient} for evaluation to better reflect real-world deployment scenarios. Instead of the CoT approach, which generates a large number of output tokens, we adopt a few-shot prompting strategy by including examples in the input tokens. Specifically, we apply the following settings: \textit{KMMLU} (5-shot), \textit{HARRAE} (5-shot), \textit{GSM8K-Ko} (5-shot), \textit{MMLU} (5-shot), and \textit{BBH} (3-shot).

\section{Telco Benchmark Details}
\label{sec:telco_benchmark_details}

The Telco dataset is primarily composed of call center-related data. Except for a telecommunications/network terminology test designed to assess telecom and network knowledge, most tasks are related to call center operations. Each task was developed based on real use cases and has been thoroughly validated by telecommunications and customer service experts. Table \ref{tab:telcoBenchDetails_1} and \ref{tab:telcoBenchDetails_2} below presents the tasks along with examples. The overall evaluation was conducted using the implementation within \textit{LM Eval Harness}\cite{evalharness2021}. To compensate for the lack of telecommunications and network knowledge assessment, we conducted evaluations using a test set for the RAG QA system, as detailed in Figure~\ref{fig:nw_agent_failure_rates} and Table~\ref{tab:data_extraction_summary}.

\section{Finance DACP Datasets}
\label{sec:finance_dacp_datasets}

The unsupervised training dataset for the financial DACP was constructed from a relatively smaller amount of data compared to the Telco domain due to limited raw financial data. Based on the available financial data, we assembled the financial DACP dataset in a ratio similar to that of the Telco DACP dataset, with the results presented in Table \ref{tab:financeDataRatio}. This enabled us to evaluate the applicability of the approach to domains beyond telecommunications, such as finance.

\begin{table}[h] \label{tab:financeDataRatio}
    \small
    \centering
    \begin{tabular}{
        >{\raggedright\arraybackslash}p{0.2\textwidth}
        >{\centering\arraybackslash}p{0.1\textwidth}
        >{\centering\arraybackslash}p{0.1\textwidth}
    }
        \toprule 
        \textbf{Dataset} &\textbf{Size} &\textbf{Ratio} \\
        \midrule
        Finance Corpora &24.5GB &46.22\% \\
        Korean Corpora &20GB &37.73\% \\
        English Corpora &5.5GB &10.37\% \\
        Code &3GB &5.66\% \\
        \midrule
        \textbf{Total} &53GB &100\% \\
        \bottomrule
    \end{tabular}
    \vspace{-1mm}
    \caption{Size and ratio of training dataset for Finance DACP.}
    \vspace{-3mm}
    \label{tab:financeDataRatio}
\end{table}

\section{Finance Benchmark Details}
\label{sec:finance_benchmark_details}


For the financial domain evaluation, we curated a benchmark dataset by selecting tasks commonly encountered in the finance industry. Subject matter experts formulated queries and answers across three categories: passage-based multiple-choice QA (\textit{FinPMCQA}, Logit Likelihood), passage-based generation QA (\textit{FinPQA}, Korean PoS Rouge-L), and document summarization (\textit{FinSM}, Korean PoS Rouge-L), ensuring a diverse set of cases for each category. These benchmarks are composed based on various financial statements, financial reports, and contractual terms. Among these, examples related to contractual terms can be found in Table \ref{tab:financeBenchDetails_1} and \ref{tab:financeBenchDetails_2}. The overall evaluation was conducted using the implementation within \textit{LM Eval Harness}\cite{evalharness2021}.

\begin{table}[t]
    \centering
    \small
    \begin{tabular}{lcc}
        \toprule
        \textbf{Models} & \textbf{hit@3} & \textbf{Finance} \\
        \midrule
        \exaone 3.5 7.8B IT & 0.91 & 0.633 \\
        \exaone 3.5 7.8B + RAG FT & 0.91 & 0.666 \\
        \rowcolor[gray]{0.9} \exaone Finance 7.8B + RAG FT & 0.91 & \textbf{0.683} \\
        \bottomrule
    \end{tabular}
    \caption{Comparison of Model Performance in Finance Domain}
    \label{tab:fin_rag_model_performance}
\end{table}

\begin{table*}[h] \label{tab:TaskExplain}
    \footnotesize 
    \centering
    \renewcommand{\arraystretch}{0.5}
    \begin{tabular}{
        >{\raggedright\arraybackslash}p{0.12\textwidth}
        >{\centering\arraybackslash}p{0.10\textwidth}
        >{\raggedright\arraybackslash}p{0.70\textwidth}
    }
        \toprule 
        \textbf{Telco Task} &\textbf{\# of Items} &\textbf{Description} \\
        \midrule
        Vocabulary &200 &Selecting the correct term for a given description (Logit Likelihood) \\
        \midrule
        Passage Chat &150 &Selecting the appropriate response to a customer query in customer service conversations (PoS Rouge-L Acc: Tokenizing generated text using Korean PoS tagging and selecting the candidate with the highest Rouge-L score) \\
        \midrule
        Single QA &100 &Selecting the correct response to a customer query without conversational context (BLEU Acc: Selecting the candidate with the highest BLEU score) \\
        \midrule
        Chat SM &100 &Summarizing customer service conversations (PoS Rouge-L: Tokenizing generated text using Korean PoS tagging and calculating Rouge-L with the reference text) \\
        \midrule
        Chat CLS &200 &Classifying customer service conversation types (BLEU Acc: Selecting the candidate with the highest BLEU score) \\
        \bottomrule
    \end{tabular}
    \vspace{-2mm}
    \caption{Types and descriptions of benchmarks built to evaluate Telco domain performance.}
    \vspace{-3mm}
    \label{tab:TaskExplain}
\end{table*}

\begin{table*}[t]
    \centering
    \footnotesize 
    \renewcommand{\arraystretch}{1.1}
    \begin{tabular}{
        >{\raggedright\arraybackslash}p{0.23\textwidth}
        >{\centering\arraybackslash}p{0.11\textwidth}
        >{\centering\arraybackslash}p{0.11\textwidth}
        >{\centering\arraybackslash}p{0.11\textwidth}
        >{\centering\arraybackslash}p{0.13\textwidth}
        >{\centering\arraybackslash}p{0.13\textwidth}
    }
        \toprule
        \textbf{Model} & \textbf{FinPMCQA} & \textbf{FinPQA} & \textbf{FinSM} & \textbf{Finance Avg.} & \textbf{General Avg.}  \\
        \midrule
        \multicolumn{2}{l}{\textbf{Original Instruction Model (Vanilla)}} & & &\\
        \exaone 3.5 2.4B IT  & 59.50 & 28.44 & 30.61 & 39.52 & 52.64 \\
        \hdashline
        Llama 3.1 8B IT   & 70.25 & 41.51 & 30.44 & 47.40 & 59.43\\
        Qwen 2.5 7B IT  & 61.00 & 42.02 & 37.92 & 46.98 & 60.74 \\
        \exaone 3.5 7.8B IT  & 62.50 & 29.65 & 44.45 & 45.53 & 60.10 \\
        \hdashline
        Qwen 2.5 32B IT   & 60.50 & 38.95 & 51.99 & 50.48 & 58.48 \\
        \exaone 3.5 32B IT  & 69.75 & 32.32 & 50.03 & 50.70  & 68.15\\
        \hdashline
        Llama 3.3 70B IT   & 68.75 & 48.84 & 39.10 & 52.23 & 66.05\\
        Qwen 2.5 72B IT   & 57.25 & 41.77 & 58.84 & 52.62 & 65.37\\
        \midrule
        \multicolumn{3}{l}{\textbf{Finance DACP Model (Ours)}}  & & & \\
        \exaone Finance 2.4B IT & \textbf{60.25} & \textbf{62.31} & \textbf{54.08} & \textbf{58.88(+49\%)} & 49.64(-6\%) \\
        \hdashline
        \exaone Finance 7.8B IT  & \textbf{63.00} & \textbf{56.91} & \textbf{58.71} & \textbf{59.54(+31\%)} & 59.56(-1\%)  \\
        
        \bottomrule
    \end{tabular}
    \vspace{-1mm}
    \caption{Performance comparison of original instruction models and Finance Domain Adapted models (Finance DACP) evaluated on the Finance and General Domain Benchmarks. Models are grouped by size (dashed lines). \textbf{Bold} scores indicate an improvement over the instruction model. '\textit{FinPMCQA}', '\textit{FinPQA}', and '\textit{FinSM}' denote Finance Passage-based Multiple Choice Question Answering, Finance Passage-based Question Answering, and Finance Summarization tasks, respectively. Percentage improvements over original models are shown in parentheses. DACP was applied only to \exaone 3.5 2.4B and 7.8B models. }
    \vspace{-4mm}
    \label{tab:financeBench}
\end{table*}

\section{Call Center Application Evaluation Details}
\label{sec:industry_application_evaluation_details}


To evaluate the effectiveness of the DACP-applied model in a Call Center scenario, we conducted a human evaluation comparing two models: the baseline model and Telco LLM. Table~\ref{tab:evaluation_criteria} presents the evaluation criteria, where each summarization task has a distinct objective, and annotators assign scores ranging from 1 to 5. A bilingual dialogue used in the evaluation—corresponding to Figure 3—is provided in Table~\ref{tab:figure3_conversation_translation}.

\section{RAG System in Finance Evaluation Details}

Table \ref{tab:fin_rag_model_performance} presents the overall performance of the financial RAG system from the leaderboard in \cite{allganize2024rag}. With top-3 passages, the DACP-applied model outperforms the baseline models.


Table \ref{tab:human_evaluation_framework} presents the evaluation criteria and metrics for each category, with rank ranging from 1 to 5.

\begin{table*}
    \small
    \centering
    \begin{tabular}{p{3cm} p{5.5cm} p{5.5cm}}
        \toprule 
        \textbf{Evaluation Criteria} & \textbf{Customer Request Summary} & \textbf{Agent's Response Summary} \\
        \midrule
        Score Range & 1 - 5 & 1 - 5 \\
        \midrule
        Evaluation Objective & Assessing the extent to which the customer's request is summarized (i.e., how well the summary allows for historical tracking in future references). & Assessing the clarity of the agent's response summary in conveying the resolution of the request (i.e., how well the summary allows for historical tracking in future references). \\
        \midrule
        Evaluation Standards & \textbf{1}: The summary does not convey the customer's request at all. 
        
        \textbf{3}: The summary partially conveys the customer's request, making about half of it understandable.
        
        \textbf{5}: The summary is concisely written as if composed by an agent, allowing for effective historical tracking. & \textbf{1}: The summary does not convey the resolution provided to the customer at all.
        
        \textbf{3}: The summary partially conveys the resolution, making about half of it understandable.
        
        \textbf{5}: The summary is concisely written as if composed by an agent, allowing for effective historical tracking. \\
        \bottomrule
    \end{tabular}
    \vspace{-1mm}
    \caption{Evaluation framework for measuring the quality of customer request and agent response summaries for historical tracking purposes}
    \vspace{-3mm}
    \label{tab:evaluation_criteria}
\end{table*}

\begin{table*}[h]
    \small
    \centering
    \begin{tabular}{p{2.4cm} p{6.0cm} p{6.0cm}}
        \toprule
        \textbf{Role} & \textbf{Original (KR)} & \textbf{Machine Translated (EN)} \\
        \midrule
        User ① & ES727장비에서 filter dhcpv6 gi1-24는 무슨 뜻인가요? & In the ES727 device, what does ``filter dhcpv6 gi1-24'' mean? \\
        \midrule
        Assistant ① & ‘filter dhcpv6 gi1-24’는 ES727AR 장비에서 DHCPv6 프로토콜을 사용할 때, 포트 범위(gi1-24)에 인입되는 reply 및 advertise 패킷을 차단하는 설정입니다. 주로 가입자 모드에 적용되어 불필요한 DHCPv6 트래픽을 제어할 수 있습니다. & ``filter dhcpv6 gi1-24'' is a setting on the ES727AR device that blocks reply and advertise packets coming into the port range gi1-24 when using the DHCPv6 protocol. It is typically applied in subscriber mode and helps suppress unnecessary DHCPv6 traffic. \\
        \midrule
        User ② & LPS-4100 사용 설명서 찾아줘 & Please find the LPS-4100 user manual for me. \\
        \midrule
        Assistant ② & LPS-4100 사용 설명서를 찾으시는군요. 관련 문서: 장비매뉴얼\_국사시설기술팀\_오재현\_LPS-4100\_사용\_설명서\_v1.1.pdf (17--21쪽, 설치·설정 방법 포함). & Here’s a document that contains information on the LPS-4100: Equipment-Manual-Network-Facilities-Team-OJaehyun-LPS-4100-User-Guide-v1.1.pdf (pp.~17--21; installation and configuration details). \\
        \midrule
        User ③ & 고마워, 오늘 뭐 먹을지 고민되는데 점심 메뉴 추천해줄래? & Thanks! I’m wondering what to eat today—could you recommend a lunch menu? \\
        \midrule
        Assistant ③ & 점심 메뉴로는 다양한 선택지가 있습니다. 몇 가지 추천해드릴게요: 1.~\textbf{비빔밥}: 신선한 채소와 고기를 함께 비벼 먹는 건강한 한 끼입니다. 2.~\textbf{김치찌개}: 얼큰하고 따뜻한 국물이 일품인 김치찌개는 밥과 함께 먹기 좋습니다. 3.~\textbf{불고기 덮밥}: 달콤하고 짭짤한 불고기를 밥 위에 얹어 먹는 간편한 메뉴입니다. 오늘의 기분이나 날씨에 따라 선택해보세요! & There are plenty of lunch options. Here are a few suggestions: 1.~\textbf{Bibimbap}: A healthy bowl mixing fresh vegetables and meat with rice. 2.~\textbf{Kimchi stew}: A spicy, warming soup that pairs well with rice. 3.~\textbf{Bulgogi rice bowl}: Sweet and savory bulgogi served over rice for an easy meal. Choose according to your mood or today’s weather! \\
        \bottomrule
    \end{tabular}
    \caption{Example bilingual conversation between a customer (User) and service agent (Assistant).}
    \label{tab:figure3_conversation_translation}
\end{table*}

\begin{table*}[h]
    \small
    \centering
    \begin{tabular}{p{5cm} c  p{4cm} p{4cm}}
        \toprule
        \textbf{Evaluation Category} & \textbf{Scale} & \textbf{Evaluation Criteria} & \textbf{Evaluation Basis} \\
        \midrule
        Data Extraction, Document Understanding & 1 -- 5 & Level of accurate data extraction from the document & MRR: Mean reciprocal rank \\
        \midrule
        Document Organization based on Freshness, Format Identification \& Recommendation & 1 -- 5 & Level of freshness recognition and formatting capability & MRR: Mean reciprocal rank \\
        \midrule
        Document Generation & 1 -- 5 & Level of document generation capability & MRR: Mean reciprocal rank \\
        \midrule
        Memorization of Content & 1 -- 5 & Level of content memorization using documents & MRR: Mean reciprocal rank \\
        \bottomrule
    \end{tabular}
    \caption{The table below describes evaluation and presents a systematic framework for assessing machine reading comprehension capabilities in the finance domain. Each evaluation category is measured on a 1-5 ranking scale and quantified using the MRR (Mean Reciprocal Rank) methodology.}
    \label{tab:human_evaluation_framework}
\end{table*}

\begin{table*}[t] \label{tab:telcoBenchDetails_1}
    \small
    \centering
    \begin{tabular}{
    >{\raggedright\arraybackslash}p{0.10\textwidth}
    >{\centering\arraybackslash}p{0.09\textwidth}
    >{\raggedright\arraybackslash}p{0.33\textwidth}
    >{\raggedright\arraybackslash}p{0.37\textwidth}
}
    \toprule 
    \textbf{Telco Task} &\textbf{\# of Items} &\textbf{Original Sample} & \textbf{Sample (Machine Translated)}\\
    \midrule
    \tiny{Vocabulary} &\tiny{200} & \tiny{다음 중 "통신기기설치불량"이 의미하는 내용을 잘 설명한 것은?} &
\tiny{Which of the following best explains the meaning of “Improper Installation of Communication Equipment”?} \\
& & \tiny{① 고객 요구와 기호에 부합하는 서비스를 일대일로 전개하는 마케팅 기법} &
\tiny{① A marketing method that provides one-on-one services tailored to customer preferences.} \\
& & \tiny{② 데이터 전송 시 일정한 시간폭(Time slot)으로 나누어 가입자에게 차례로 분배/전송하여, 여러 가입자 또는 데이터가 하나의 고속 전송로를 함께 공유하는 기술} &
\tiny{② A technology that divides data transmission into time slots and distributes/transmits them sequentially to subscribers, allowing multiple subscribers or data streams to share a high-speed transmission line.} \\
& & \tiny{③ 한전공가용역순시 개선요구사항(12가지 유형) 중 하나이며, 통신기기가 조가선에 고정이 불량한 상태} & 
\tiny{③ One of 12 categories of improvement requirements for KEPCO utility pole service inspections, indicating that communication equipment is improperly fixed to the support wire.} \\
& & \tiny{④ NHN엔터테인먼트의 간편결제 서비스명  ※ 간편결제: 최초 1회 회원가입 및 결제정보(신용카드번호/은행계좌번호 등) 등록 후, 결제정보 입력 없이 간단한 인증만으로 결제} &
\tiny{④ NHN Entertainment's simple payment service. ※ Simple payment: After registering payment information (credit card/bank account) once, users can make payments with simple authentication without re-entering payment details.} \\
    \midrule
    \tiny{Passage Chat} &\tiny{150} & \tiny{위 정보를 참고하여 아래 대화의 USER의 질문에 가장 적절한 답을 생성하세요} &
\tiny{Please generate the most appropriate response to the USER’s question based on the following information.} \\
& & \tiny{\textbf{요금제 테이블 : }} &
\tiny{\textbf{Plan Table:}} \\
& & \tiny{5G 시그니처 | 130,000원 | 무제한 데이터 | 120GB 공유 데이터 } &
\tiny{5G Signature | 130,000 KRW | Unlimited Data | 120GB sharing data} \\
& & \tiny{5G 프리미어 플러스 | 105,000원 | 무제한 데이터 | 100GB 공유 데이터 } &
\tiny{5G Premier Plus | 105,000 KRW | Unlimited Data | 100GB sharing data} \\
& & \tiny{5G 프리미어 에센셜 | 85,000원 | 무제한 데이터 | 30GB 공유 데이터} &
\tiny{5G Premier Essential | 85,000 KRW | Unlimited Data | 30GB sharing data} \\
& & \tiny{5G 스탠다드 | 75,000원 | 150GB 데이터 | 10GB 공유 데이터} &
\tiny{5G Standard | 75,000 KRW | 150GB Data | 10GB sharing data} \\
& & \tiny{5G 데이터 플러스 | 66,000원 | 80GB 데이터 | 10GB 공유 데이터} &
\tiny{5G Data Plus | 66,000 KRW | 80GB Data | 10GB sharing data} \\
& & \tiny{* 5G 시그니처 요금제와 5G 프리미어 플러스 요금제는 프리미엄혜택 1개 무료 제공(프리미엄혜택: 티빙, 디즈니+, 넷플릭스 등)} &
\tiny{* Other plans vary. Premium plans offer 1 free benefit (e.g., TVING, Disney+, Netflix, etc.).} \\
& & \tiny{\textbf{고객 사용량 (3개월 평균):}} &
\tiny{\textbf{Customer Usage (3-month average):}} \\
& & \tiny{5G 데이터 플러스 요금제 사용중, 데이터 제공량 80GB 중 평균 150GB 사용, 쉐어링 데이터 제공량 10GB 중 평균 5GB 사용} &
\tiny{Using 5G Data Plus. Average data use: 150GB (exceeds 80GB limit). Average sharing data use: 5GB (within 10GB limit).} \\
& & \tiny{\textbf{대화:}} &
\tiny{\textbf{Conversation:}} \\
& & \tiny{C: 안녕하세요! 요즘 데이터를 너무 많이 쓰는 것 같아서, 요금제를 바꿀까 해요. 어떤 걸로 바꾸면 좋을까요?} &
\tiny{C: Hi! I’ve been using a lot of data lately and I’m thinking about changing my plan. Any suggestions?} \\
& & \tiny{A: 안녕하세요! 고객님의 사용량을 확인해보니, 평균 데이터 사용량이 150GB이고 쉐어링 데이터 사용량이 5GB입니다. 현재 사용 중인 5G 데이터 플러스 요금제의 데이터 제공량을 초과하고 있습니다. 이에 따라 5G 프리미어 에센셜 요금제를 추천드립니다. 이 요금제는 무제한 데이터를 제공하며, 쉐어링 데이터도 30GB를 제공합니다. 월 85,000원입니다.} &
\tiny{A: Hello! Based on your usage, you’re averaging 150GB of data per month, which exceeds your current plan’s 80GB limit. I recommend the 5G Premier Essential plan. It offers unlimited data and 30GB of sharing data for 85,000 KRW per month.} \\
& & \tiny{C: 데이터 무제한이라면 초과 요금 걱정은 없겠네요. 그런데 요금이 조금 더 저렴한 옵션은 없을까요?} &
\tiny{C: Unlimited data sounds great—no more overage fees! Is there a cheaper option?} \\
& & \tiny{\textbf{선택지:}} &
\tiny{\textbf{Candidate Responses:}} \\
& & \tiny{① 5G 프리미어 에센셜 요금제보다 저렴한 요금제로는 5G 스탠다드 요금제가 있습니다. 이 요금제는 월 75,000원에 150GB의 데이터와 10GB의 쉐어링 데이터를 제공합니다. 다만, 고객님의 평균 사용량을 고려했을 때 데이터 초과 사용으로 인해 추가 요금이 발생할 수 있습니다.} &
\tiny{① The 5G Standard plan offers 150GB data and 10GB sharing for 75,000 KRW, but overage charges may apply based on your usage.} \\
& & \tiny{② 5G 프리미어 에센셜보다 저렴한 요금제로는 5G 라이트+가 있습니다. 이 요금제는 월 55,000원에 12GB의 데이터와 20GB의 쉐어링 데이터를 제공합니다.} &
\tiny{② The 5G Light+ plan offers 12GB data and 20GB sharing for 55,000 KRW, but may not suit your needs.} \\
& & \tiny{③ 5G 데이터 슈퍼 요금제가 있습니다. 이 요금제는 월 68,000원에 95GB의 데이터를 제공하며, 쉐어링 데이터는 15GB가 제공됩니다.} &
\tiny{③ The 5G Data Super plan offers 95GB data and 15GB sharing for 68,000 KRW.} \\
& & \tiny{④ 데이터 무제한 요금제는 정말 편리하죠. 혹시 최근에 출시된 스마트폰에 대해 관심 있으신가요} &
\tiny{④ Unlimited data is convenient! Are you interested in the latest smartphones?} \\
& & \tiny{⑤ 저렴한 요금제도 좋지만, 여행을 계획 중이시라면 여행 일정도 함께 고려해 보는 것이 좋습니다. 요즘 인기 있는 여행지에 대해 알려드릴까요} &
\tiny{⑤ Lower plans are great, but if you’re planning a trip, we can explore travel-friendly options. Would you like recommendations on trending destinations?} \\
    \midrule
    \tiny{Single QA} & \tiny{100} & \tiny{다음은 상담 대화 중 사용자의 질문입니다. 올바른 답변을 선택하세요.} & \tiny{The following is a customer question during a service conversation. Choose the correct response.} \\
& & \tiny{Q: 요금제 프리미엄 혜택 중에 뭐가 있나요?} & \tiny{Q: What benefits are included in the premium plan? } \\
& & \tiny{① 프리미엄 회원 혜택 중에는 무제한 커피 리필 서비스가 제공됩니다. 이는 최근 실시된 커피 관련 연구에서 커피를 무제한으로 마시는 것이 건강에 좋다는 연구 결과에 근거하고 있습니다.} &
\tiny{① One of the benefits for premium members is unlimited coffee refills. This is based on recent research suggesting that unlimited coffee consumption is beneficial to health.} \\
& & \tiny{②프리미엄 혜택으로는 다음 중 하나를 선택하실 수 있습니다: 티빙, 디즈니+, 넷플릭스(베이식)+유튜브 프리미엄, 넷플릭스(스탠다드), 헬로렌탈구독, 일리커피구독, 구글 패키지입니다. 이 중에서 원하시는 혜택을 선택하여 무료로 이용하실 수 있습니다.} & \tiny{② As a premium benefit, you may choose one of the following services: TVING, Disney+, Netflix (Basic) + YouTube Premium, Netflix (Standard), Hello Rental Subscription, Illy Coffee Subscription, or Google Package. You may enjoy the selected benefit for free}. \\
& & \tiny{③ 죄송합니다. 질문하신 내용에 대해 알려드릴 수 있는 정보가 없습니다. 추가 질문이 있으시다면 도와드리겠습니다.} & 
\tiny{③ Sorry, we are unable to provide information regarding your question. Please feel free to ask additional questions.} \\
& & \tiny{④ 요금제 변경은 매월 첫째 주에만 가능합니다. 고객센터로 전화하시거나 요금제 페이지에서 변경할 수 있습니다. } & 
\tiny{④ Plan changes are only allowed during the first week of each month. You may contact customer service or change your plan via the plan management page.}\\
\bottomrule
\end{tabular}
    \vspace{-1mm}
    \caption{Telco Benchmark Examples (1)}
    \vspace{-3mm}
    \label{tab:telcoBenchDetails_1}
\end{table*}

\begin{table*}[t] \label{tab:telcoBenchDetails_2}
    \small
    \centering
    \begin{tiny}
    \begin{tabular}{
    >{\raggedright\arraybackslash}p{0.1\textwidth}
    >{\centering\arraybackslash}p{0.09\textwidth}
    >{\raggedright\arraybackslash}p{0.33\textwidth}
    >{\raggedright\arraybackslash}p{0.37\textwidth}
}
    \toprule 
    \small{\textbf{Telco Task}} & \small{\textbf{\# of Items}} &\small{\textbf{Original Sample}} & \small{\textbf{Sample (Machine Translated)}} \\
    \midrule
    Chat SM &100 & 다음 상담 대화를 요약하세요. &
Summarize the following customer service conversation. \\ \\
& & \textbf{대화: } & 
\textbf{Conversation:} \\
& & A: 안녕하세요, 통신사입니다. <NAME> 고객님 맞으신가요? & A: Hello, this is Telecom. Are you <NAME>? \\ 
& & C: 네, 맞습니다. 자동이체를 삼성카드에서 현대카드로 바꾸고 싶어요. & C: Yes, I am. I want to change my auto-payment from Samsung Card to Hyundai Card. \\ 
& & A: 네, 변경할 카드가 고객님 본인 카드 맞으신가요? 생년월일과 카드번호, 유효기간 말씀 부탁드립니다. & A: Is the new card in your name? Please provide your date of birth, card number, and expiry date. \\ 
& & C: 네, <BIRTH\_NUMBER>, <CARD\_NUMBER>, 유효기간은 <DATE>입니다. & C: Yes, <BIRTH\_NUMBER>, <CARD\_NUMBER>, expiry date is <DATE>. \\ 
& & A: 확인 감사합니다. <DATE>부터 현대카드로 요금 결제가 시작됩니다. 혹시 다른 가족분 번호도 변경 필요하신가요? & A: Thank you. Your payments will be charged to Hyundai Card starting from <DATE>. Do you need to change any family members’ numbers as well? \\ 
& & C: 아니요, 없습니다. & C: No, I don't. \\ 
& & A: 추가로 궁금한 점 있으신가요?& A: Do you have any other questions? \\ 
& & C: 제가 VIP인데 혜택은 없나요? & C: I’m a VIP. Are there any benefits? \\
& & A: 고객님은 VVIP로 조회되며, GS25, 파리바게뜨에서 멤버십 바코드로 할인받으실 수 있습니다. 혜택 안내를 문자로 보내드릴까요? & A:You are registered as VVIP. You can get discounts at GS25 and Paris Baguette using your membership barcode. Shall I send the benefit info by text? \\ 
& & C: 네, 감사합니다. & C: Yes, thank you. \\ 
& & A: 네, 카드 자동이체 변경 완료되었습니다. 좋은 하루 보내세요. & A: Your auto-payment change has been completed. Have a great day. \\ 
    \midrule
    Chat CLS &200 & 다음 상담 대화를 보고, 상담 유형을 분류하세요. & Read the following customer service conversation and classify the type of inquiry. \\ \\
& & \textbf{대화:} & \textbf{Conversation:} \\
& & A: 안녕하세요, U+입니다. <NAME> 고객님 맞으신가요? & A: Hello, this is Telecom. Are you <NAME>? \\
& & C: 네, 맞습니다. 엄마 카드로 자동이체 되는데 카드가 고장 나서 재발급 받아야 해요. & C: Yes. Auto-pay is set to my mom’s card, but it’s broken and needs to be reissued. \\
& & A: 네, 그런 경우에는 자동이체를 새 카드로 변경해야 진행됩니다. 카드 재발급 받으셨나요? & A: In that case, auto-pay needs to be changed to the new card. Have you received the reissued card? \\
& & C: 아니요, 받기 전에 문의드리려고 전화한 거예요. & C: Not yet, I called to ask before receiving it. \\
& & A: 네, 재발급 받으시면 저희에게 연락 주셔서 자동이체 변경을 진행해 주시면 됩니다. & A: Once you receive it, please contact us again to update the auto-payment. \\
& & C: 알겠습니다. <DATE> 안으로 처리해야 요금이 빠져나가는 건가요? & C: Got it. Do I need to take care of this before <DATE> to avoid missing the payment? \\
& & A: 네, <DATE>까지 결제하시면 미납이 되지 않으니 걱정하지 않으셔도 됩니다. 다른 궁금한 사항 있으신가요? & A: Yes, as long as payment is made by <DATE>, there will be no overdue charges. Any other questions? \\
& & C: 5G 요금제 변경은 상관 없나요? & C: Is it okay if I change my 5G plan? \\
& & A: 네, 현재 최저 요금제로 사용 중이시니 문제 없습니다. 다른 문의 사항 있으면 연락 주세요. & A: Yes, you are currently on the lowest plan, so there’s no issue. Feel free to contact us if you have more questions. \\ \\
& & \textbf{선택지:} & \textbf{Candidates:} \\
& & ① 잘못걸린 전화 (내부) & ① Wrong Number (Internal) \\
& & ② 청구서 용어 및 당월 청구요금 문의 & ② Billing Terms and Current Charges Inquiry \\
& & ③ 자동이체 방법 변경/문의 & ③ Auto\-payment Change/Inquiry \\
& & ④ 홈 요금제 위약금/약정문의 & ④ Penalty/Contract Inquiry of Home Plan \\
\bottomrule
\end{tabular}
\end{tiny}
    \vspace{-1mm}
    \caption{Telco Benchmark Examples (2)}
    \vspace{-3mm}
    \label{tab:telcoBenchDetails_2}
\end{table*}

\begin{table*}[t] \label{tab:financeBenchDetails_1}
    \small
    \centering
    \begin{tiny}
    \begin{tabular}{
    >{\raggedright\arraybackslash}p{0.08\textwidth}
    >{\centering\arraybackslash}p{0.09\textwidth}
    >{\raggedright\arraybackslash}p{0.33\textwidth}
    >{\raggedright\arraybackslash}p{0.38\textwidth}
}
    \toprule 
    \small{\textbf{Fin. Task}} &\small{\textbf{\# of Items}} &\small{\textbf{Original Sample}} &\small{\textbf{Sample (Machine Translated)}} \\
    \midrule
FinPMCQA  &400 & 본문을 읽고 아래 질문에 올바른 답을 고르면? & Read the passage and choose the correct answer to the question below. \\ \\ 
& & \textbf{[본문]} & \textbf{[Passage]}\\ 
& & 제3조(서비스 이용계약) & Article 3 (Service Usage Agreement)\\ 
& & (1) 휴대폰 메시지 서비스를 이용하려는 자는 카드회원 가입 시 서비스 이용을 신청하거나 회사 전화, 홈페이지 등을 통해 가입을 신청해야 하고 유효한 휴대폰 번호를 회사에 제공해야 합니다. & (1) A person who wishes to use the mobile message service must apply at the time of card membership registration or through the Company’s telephone service, website, etc., and must provide a valid mobile phone number to the Company.\\ 
& & (2) 휴대폰 메시지 서비스는 연중무휴, 1일 24시간 제공되며 회사 시스템의 유지‧보수‧점검(업무 위‧수탁 업체를 포함한다), 이동통신사 또는 인터넷망 사업자의 시스템 유지‧보수‧점검 등으로 휴대폰 메시지 서비스 이용이 제한, 지연, 누락 될 경우 1개월 전에 이용자에게 이용대금명세서, 서면, 전화, 전자우편(E-mail), 휴대폰 메시지 등으로 동 내용을 안내해야 합니다. 다만, 긴급한 사정으로 휴대폰 메시지 서비스 이용이 제한, 지연, 누락된 사실을 미리 안내할 수 없는 경우에는 전달되지 못한 휴대폰 메시지 서비스 내용을 이용자에게 즉시 다시 전송해야 하며 사후적으로 휴대폰 메시지 서비스 이용이 제한, 지연, 누락되어 이용자에게 손해가 발생한 경우에는 이를 배상하도록 합니다.& (2) The mobile message service is provided 24 hours a day, 365 days a year. However, if the service is restricted, delayed, or omitted due to maintenance, repair, or inspection of the Company’s system (including subcontracted service providers), or due to the maintenance, repair, or inspection of the systems of telecom or internet network operators, the user must be notified at least one month in advance via billing statements, written notice, phone call, email, or mobile message. In cases of urgent circumstances where advance notice cannot be given, the undelivered mobile message must be resent to the user immediately, and if the user suffers any damages due to the restriction, delay, or omission of the mobile message service, the Company must compensate for those damages.\\ & & (3) 제2항 단서에도 불구하고 다음 각 호의 경우에는 이용자가 손해배상을 청구할 수 없습니다. & (3) Notwithstanding Paragraph (2), the user may not claim compensation for damages in the following cases:\\ 
& & 이용자의 휴대폰 분실, 도난, 파손, 고장 등 이용자의 책임있는 사유로 휴대폰 메시지를 정상적으로 수신할 수 없는 경우 & If the user cannot properly receive the mobile message due to reasons attributable to the user, such as loss, theft, damage, or malfunction of their mobile phone.\\ 
& & 이용자의 휴대폰 전원이 꺼져있거나, 이용자의 책임있는 사유로 휴대폰이 정지 또는 해지된 경우 & If the mobile phone is turned off or is suspended or terminated due to reasons attributable to the user.\\ 
& & 이용자의 휴대폰 번호가 변경된 사실을 회사에 알리지 않은 경우 & If the user fails to notify the Company of a change in their mobile phone number.\\ 
& & (4) 회사는 휴대폰 메시지 서비스 제공 범위에 따라 이용자별로 개인회원(본인회원 및 가족회원별) 300원의 수수료를 청구할 수 있습니다. 수수료를 변경할 경우 변경 전 3개월부터 매월 홈페이지, 이용대금명세서, 서면, 우편, 전화, 전자우편(E-mail), 휴대폰메시지 중 2가지 이상의 방법으로 이용자에게 수수료 변경 사유와 변경 내용, 수수료 변경에 따른 계약해지 방법을 안내해야 합니다. & (4) The Company may charge a fee of 300 KRW per individual member (including the main member and family members) depending on the scope of the mobile message service provided. If the fee is to be changed, starting three months before the change, the Company must notify users each month of the reason for the change, the details of the change, and the method for contract termination due to the fee change, through at least two of the following means: website, billing statement, written notice, postal mail, telephone, email, or mobile message.\\ \\ 
& & \textbf{[질문]} 휴대폰 메시지 서비스 이용 수수료는 어떻게 청구되나요? & \textbf{[Question]} How is the mobile message service fee charged? \\ 
& & ① 이용자는 수수료를 선택적으로 납부할 수 있다 & ① The user may choose whether or not to pay the fee.\\ 
& & ② 모든 이용자에게 월 300원의 수수료가 부과된다. & ② A monthly fee of 300 KRW is charged to all users.\\ 
& & ③ 이용자별로 개인회원에 한해 300원의 수수료가 청구된다. & ③ A fee of 300 KRW is charged per individual member.\\ 
& & ④ 수수료는 휴대폰 번호 변경 시에만 부과된다. & ④ The fee is only charged when the mobile phone number is changed.\\
\bottomrule
\end{tabular}
\end{tiny}
    \vspace{-1mm}
    \caption{Finance Benchmark Examples (1)}
    \vspace{-3mm}
    \label{tab:financeBenchDetails_1}
\end{table*}

\begin{table*}[t] \label{tab:financeBenchDetails_2}
    \small
    \centering
    \begin{tiny}
    \begin{tabular}{
    >{\raggedright\arraybackslash}p{0.08\textwidth}
    >{\centering\arraybackslash}p{0.09\textwidth}
    >{\raggedright\arraybackslash}p{0.33\textwidth}
    >{\raggedright\arraybackslash}p{0.38\textwidth}
}
    \toprule 
    \small{\textbf{Fin. Task}} &\small{\textbf{\# of Items}} &\small{\textbf{Original Sample}} &\small{\textbf{Sample (Machine Translated)}} \\
    \midrule
    FinSM &250 & 본문의 내용을 3문장 이하로 요약하시오. & Summarize the content of the passage in no more than three sentences. \\ \\
& & \textbf{[본문]} & \textbf{[Passage]} \\
& & 제 3 조(약관 명시 및 개정 등) & Article 3 (Stipulation and Revision of Terms and Conditions) \\
& & ① 본 약관은 “서비스”가 금융혁신지원특별법에 따른 혁신금융서비스로 선정됨에 따라 제정된 약관으로, 동 특별법이 지정하는 기간 내에서만 “서비스”제공이 가능하며, 지정 기간 종료, 금융당국의 중지 또는 변경 명령 등에 의하여 변경될 수 있습니다. & ① These terms and conditions have been established as the service has been designated as an innovative financial service under the Special Act on Financial Innovation Support, and the service can only be provided within the period specified by the Act. The service may be changed or suspended upon the expiration of the designated period or by order of the financial authorities. \\
& & ② “회사”는 이 약관의 내용과 상호, 영업소 소재지, 대표자의 성명, 사업자등록번호, 연락처(전화, 팩스, 전자우편주소 등), 개인정보관리책임자 등을 이용자가 알 수 있도록 “회사” 홈페이지에 게시합니다. 다만, 게시된 내용은 “서비스이용화면”과의 연결하여 “이용자”가 볼 수 있도록 할 수 있습니다. & ② The company shall post the contents of these terms and conditions, company name, business office location, representative’s name, business registration number, contact information (phone, fax, email address, etc.), and personal information manager on the company website for users to access. However, the posted information may be linked to the service usage screen for user access. \\
& & ③ “회사”는 금융혁신지원특별법, 여신전문금융업법, 독점규제및공정거래에관한법률 등 관련법을 위배하지 않는 범위에서 이 약관을 개정할 수 있습니다. & ③ The company may revise these terms and conditions to the extent that such revision does not violate relevant laws such as the Special Act on Financial Innovation Support, the Specialized Credit Finance Business Act, and the Monopoly Regulation and Fair Trade Act. \\
& & ④ “회사”가 이 약관을 개정할 경우에는 그 내용을 적용 예정일로부터 1 개월 이전까지 “이용자”에게 통지하며, “이용자”가 적용 예정일까지 “회사”에 계약해지 의사표시를 하지 않을 경우에는 변경된 약관을 승인한 것으로 간주합니다. 한편, “회사”는 이러한 통지 외에도 “서비스이용화면”에 현행 약관과 함께 적용일자 및 개정사유를 게시합니다. 이 경우 “회사”는 개정 전 내용과 개정 후 내용을 명확하게 비교하여 “이용자”가 알기 쉽도록 표시합니다. & ④ When revising these terms and conditions, the company shall notify users at least one month before the scheduled effective date. If the user does not express their intent to terminate the contract by the effective date, it shall be deemed that the user has agreed to the revised terms and conditions. Additionally, the company shall post the current terms and conditions along with the effective date and reason for the revision on the service usage screen, clearly comparing the changes before and after for easy understanding by users. \\
& & ⑤ “회사”가 제 4 항에 따라 약관 변경에 관하여 게시하거나 통지를 하는 경우에는 “이용자”가 약관의 변경내용이 게시되거나 통지된 후부터 변경되는 약관의 시행일 전의 영업일까지 계약을 해지할 수 있으며, 약관의 변경 내용에 이의를 제기하지 아니하는 경우 약관의 변경내용에 승인한 것으로 본다.’라는 취지의 내용을 명시하여 게시 및 통지합니다. & ⑤ When the company posts or notifies changes to the terms and conditions under Paragraph ④, it shall explicitly state that users may terminate the contract until the business day before the effective date and that failure to raise objections shall be considered as consent to the revised terms and conditions. \\

    \midrule
FinPQA  &400 & 다음 본문을 보고 질문에 답하시오.  & Read the passage and choose the correct answer to the question below. \\ \\ 
& & \textbf{[본문]} & \textbf{[Passage]}\\ 
& &  제20조(청약의 철회) & Article 20 (Withdrawal of Application)\\ 
& & ① 계약자는 보험증권을 받은 날 부터 15일 이내에 그 청약을 철회할 수 있습니다. 다만, 회사가 건강 상태 진단을 지원하는 계약, 보험기간이 90일 이내인 계약 또는 전문금융소비자가 체결한 계약은 청약을 철회할 수 없습니다. & ① The policyholder may withdraw the application within 15 days from the date of receiving the insurance policy. However, withdrawal is not permitted for contracts where the company provides a health status examination, contracts with an insurance period of 90 days or less, or contracts entered into by professional financial consumers. \\ 
& & 【전문금융소비자】 & 【Professional Financial Consumer】 \\ 
& & 보험계약에 관한 전문성, 자산규모 등에 비추어 보험계약에 따른 위험감수능력이 있는 자로서, 국가, 지방자치단체, 한국은행, 금융회사, 주권상장법인 등을 포함하며 「금융소비자 보호에 관한 법률」제2조(정의) 제9호에서 정하는 전문금융소비자를 말합니다. & Refers to an entity that has the capacity to bear risks under an insurance contract, in consideration of expertise and asset size, including national or local governments, the Bank of Korea, financial institutions, listed companies, etc., as defined under Article 2(9) of the “Act on the Protection of Financial Consumers.” \\ 
& & 【일반금융소비자】 & 【General Financial Consumer】 \\ 
& & 전문금융소비자가 아닌 계약자를 말합니다. &Refers to a policyholder who is not a professional financial consumer.\\ 
& & ② 제1항에도 불구하고 청약한 날부터 30일(단, 만 65세 이상의 계약자가 통신수단 중 전화를 이용하여 체결한 경우 45일)이 초과된 계약은 청약을 철회할 수 없습니다. & ② Notwithstanding Paragraph ①, an application may not be withdrawn if more than 30 days (or 45 days in the case where a policyholder aged 65 or older concludes the contract via telephone) have passed since the date of application.\\ 
& & ③ 청약철회는 계약자가 전화로 신청하거나, 철회의사를 표시하기 위한 서면, 전자우편, 휴대전화 문자 메시지 또는 이에 준하는 전자적 의사표시(이하 ‘서면 등’이라 합니다)를 발송한 때 효력이 발생합니다. 계약자는 서면 등을 발송한 때에 그 발송 사실을 회사에 지체없이 알려야 합니다. & ③ The withdrawal becomes effective when the policyholder applies by phone, or sends a written notice, email, mobile text message, or any equivalent electronic notification (hereinafter referred to as “written notice, etc.”). The policyholder must immediately inform the company of the fact that the written notice, etc., has been sent.\\ 
& & ④ 계약자가 청약을 철회한 때에는 회사는 청약의 철회를 접수한 날부터 3영업일 이내에 납입한 보험료를 계약자에게 돌려드리며, 보험료 반환이 늦어진 기간에 대하여는 ‘보험개발원이 공시하는 보험계약대출이율’을 연단위 복리로 계산한 금액을 더하여 지급합니다. 다만, 계약자가 제1회 보험료 등을 신용카드로 납입한 계약의 청약을 철회하는 경우에 회사는 청약의 철회를 접수한 날부터 3영업일 이내에 해당 신용카드회사로 하여금 대금청구를 하지 않도록 해야 하며, 이 경우 회사는 보험료를 반환한 것으로 봅니다. & ④ When the policyholder withdraws the application, the company shall return the paid insurance premium within 3 business days from the date the withdrawal was received. For any delay in returning the premium, an additional amount calculated using compound annual interest based on the policy loan interest rate disclosed by the Insurance Development Institute shall be paid. However, if the first premium, etc., was paid by credit card, the company shall instruct the credit card company not to bill the amount within 3 business days from receiving the withdrawal, and in this case, the premium shall be deemed refunded. \\
& & ⑤ 청약을 철회할 때에 이미 보험금 지급사유가 발생하였으나 계약자가 그 보험금 지급사유가 발생한 사실을 알지 못한 경우에는 청약철회의 효력은 발생하지 않습니다. & ⑤ If a reason for insurance benefit payment has already occurred at the time of withdrawal, but the policyholder was unaware of such occurrence, the withdrawal shall not be effective. \\
& & ⑥ 제1항에서 보험증권을 받은 날에 대한 다툼이 발생한 경우 회사가 이를 증명하여야 합니다.  & ⑥ In the event of a dispute over the date of receipt of the insurance policy as stated in Paragraph ①, the company must provide proof. \\ \\
& & \textbf{[질문]} 청약 철회 시 납입한 보험료를 돌려주는 거야? & \textbf{[Question]} HWill the paid insurance premium be refunded upon withdrawal of application? \\ 

\bottomrule
\end{tabular}
\end{tiny}
    \vspace{-1mm}
    \caption{Finance Benchmark Examples (2)}
    \vspace{-3mm}
    \label{tab:financeBenchDetails_2}
\end{table*}

\begin{table*}
    \scriptsize
    \centering
    \begin{tabular}{p{1.7cm} p{1.3cm} p{5.5cm} p{5.5cm}}
        \toprule 
        \small{\textbf{Task}} & \small{\textbf{\# of Items}} & \small{\textbf{Original Sample}} & \small{\textbf{Sample (Machine Translated)}} \\
        \midrule
        Data Extraction and Explanation & 379 & 참조문서: 원격지원\_상품소개자료 \newline
        \newline
        내용: \newline
        4. 영업지원인력의 방문 지원은 가능한가요? \newline
        - 수도권은 방문지원 가능하며, 비수도권은 차량, KTX 등 픽업 여부에 따라 상이 하나 하반기부터는 전 지역 방문을 검토 중입니다. \newline
        - 유선 상담은 상시 가능합니다. \newline
        \newline
        5. 가입 진행 프로세스는 어떻게 되나요? \newline
        - 가입신청서 접수를 고객센터나 영업지원인력에 전달해 주시면 됩니다. \newline
        - 영업일 기준 3일 내에 오픈 가능합니다. \newline
        \newline
        6. 고객사 방문교육이 가능한가요? \newline
        - 영업지원인력과 사전 협의가 필요합니다. \newline
        \newline
        7. 기타 원격지원 판매 시 어떤 점을 강조하면 되나요? \newline
        - 원격지원은 모바일 시장 확대에 따라 PC뿐 아니라 모바일의 모든 환경(Android, iOS)과 기기에서 지원이 가능해야 합니다. \newline
        - PC는 Windows, Mobile은 Android, iOS 환경과 표준 브라우저를 지원하고 있습니다. \newline
        - 원격접속/제어만이 전부가 아닙니다. 단순히 원격제어만으로는 고객을 만족시킬 수 없습니다. \newline
        - 원격지원은 전문성이 필요한 지원업무 수행 시 동료와 협업하기(세션공유), 기록이 필요한 상담인 경우 화면녹화/상담기록 남기기 기능 등을 제공하여 고객이 만족할 수 있는 지원업무를 제공할 수 있습니다. \newline
        \newline
        문서명: 원격지원\_상품소개자료 \newline
        \newline
        End of Document \newline
        고객센터: xxxx – xxxx \newline
        문의메일: xxxx@xxxx.xx.xx \newline
        \newline
        질문: 원격지원 상품에 대한 가입 프로세스는 어떻게 진행되는지 알고 싶습니다. 또한, 영업지원인력의 방문 지원이 가능한 지역은 어디인가요? & 
        Reference Document: Remote Support Product Introduction Document \newline
        \newline
        Content: \newline
        4. Is on-site support available? \newline
        - Available in metropolitan areas. \newline
        - Availability in non-metropolitan areas depends on pickup options (e.g., vehicle, KTX). \newline
        - Nationwide on-site support is under review for the second half of the year. \newline
        - Phone consultation is always available. \newline
        \newline
        5. What is the subscription process? \newline
        - Submit the application form to the customer center or sales support staff. \newline
        - Service will be available within 3 business days. \newline
        \newline
        6. Is on-site training for clients available? \newline
        - Requires prior consultation with sales support staff. \newline
        \newline
        7. Key selling points of remote support: \newline
        - Supports all environments (PC, Android, iOS). \newline
        - Provides session sharing, screen recording, and consultation history features to enhance customer satisfaction. \newline
        \newline
        End of Document \newline
        Customer Center: xxxx – xxxx \newline
        Inquiry Email: xxxx@xxxx.xx.xx \newline
        \newline
        Question: What is the subscription process for the remote support service? Also, in which areas is on-site support available? \\
        \bottomrule
    \end{tabular}
    \caption{Each task and examples from the evaluation dataset. The dataset consists of 379 queries for which responses were generated by an LLM. A rule-based process was applied to measure the failure rate of the answers.}
    \label{tab:data_extraction_summary}
    \vspace{-1mm}
\end{table*}

\begin{table*}[h]
    \small
    \centering
    \begin{tabular}{p{5.5cm} c p{4cm} p{3.5cm}}
        \toprule
        \textbf{Evaluation Category} & \textbf{Scale} & \textbf{Evaluation Criteria} & \textbf{Evaluation Basis} \\
        \midrule
        Data Extraction, Document Understanding & 1 -- 5 & Level of accurate data extraction from the document & MRR: Mean reciprocal rank \\
        Document Organization based on Freshness, Format Identification \& Recommendation & 1 -- 5 & Level of freshness recognition and formatting capability & MRR: Mean reciprocal rank \\
        Document Generation & 1 -- 5 & Level of document generation capability & MRR: Mean reciprocal rank \\
        Memorization of Content & 1 -- 5 & Level of content memorization using documents & MRR: Mean reciprocal rank \\
        \bottomrule
    \end{tabular}
    \caption{Evaluation Framework}
    \label{tab:evaluation_framework}
\end{table*}


\end{document}